\pdfoutput=1

\documentclass[11pt]{article}

\usepackage{acl}

\usepackage{times}
\usepackage{latexsym}
\usepackage{float}

\usepackage[T1]{fontenc}

\usepackage[utf8]{inputenc}

\usepackage{microtype}

\usepackage{inconsolata}

\usepackage{url}
\usepackage{booktabs}
\usepackage{multirow}
\usepackage{makecell}
\usepackage{graphicx}
\usepackage{pifont}
\usepackage{rotating}
\usepackage{threeparttable}
\usepackage{longtable}
\usepackage{subcaption}
\usepackage{tabularx}
\usepackage{wrapfig}
\usepackage{amsmath}
\usepackage{amsfonts} 
\usepackage{amssymb}
\usepackage{arabtex}
\usepackage{utf8}
\usepackage{placeins}

\usepackage{setspace}

\usepackage{breakurl}
\usepackage[breaklinks]{hyperref}
\setcode{utf8}

\title{AceGPT, Localizing Large Language Models in Arabic}

\author{Huang Huang$^\dagger$$^{1}$, Fei Yu$^\dagger$$^2$, Jianqing Zhu$^\dagger$$^3$, Xuening Sun$^{2,4}$, Hao Cheng$^2$, Dingjie Song$^{2,4}$,\\ \bf Zhihong Chen$^{2,4}$, Abdulmohsen Alharthi$^{3}$, Bang An$^{3}$, Juncai He$^{3}$,Ziche Liu$^2$, Zhiyi Zhang$^2$,\\\bf Junying Chen$^{2,4}$, Jianquan Li$^2$, Benyou Wang\thanks{Benyou Wang is the corresponding author. \\ $^\dagger$ indicates that the three authors contributed to this work equally.}$^{2,4}$, Lian Zhang$^{1}$, Ruoyu Sun$^{1,2}$,\\ \bf Xiang Wan$^{4}$, Haizhou Li$^{2,4}$, Jinchao Xu$^{3}$\\
$^1$ Shenzhen International Center for Industrial  and  Applied  Mathematics, Shenzhen Research Institute of Big Data\\
$^2$ The Chinese University of Hong Kong, Shenzhen, China\\
$^3$ King Abdullah University of Science and Technology, Thuwal, Saudi Arabia\\
$^4$ Shenzhen Research Institute of Big Data, Shenzhen, China\\
	\texttt{wangbenyou@cuhk.edu.cn} \\
	}

\begin{document}

\maketitle

\begin{abstract}
This paper is devoted to the development of a localized Large Language Model (LLM) specifically for Arabic, a language imbued with unique cultural characteristics inadequately addressed by current mainstream models. Significant concerns emerge when addressing cultural sensitivity and local values. To address this, the paper proposes a comprehensive solution that includes further pre-training with Arabic texts, Supervised Fine-Tuning (SFT) utilizing native Arabic instructions, and GPT-4 responses in Arabic, alongside Reinforcement Learning with AI Feedback (RLAIF) employing a reward model attuned to local culture and values. The goal is to cultivate culturally cognizant and value-aligned Arabic LLMs capable of accommodating the diverse, application-specific needs of Arabic-speaking communities.
 Comprehensive evaluations reveal that the resulting model, dubbed `AceGPT', sets the state-of-the-art standard for open Arabic LLMs across various benchmarks. Codes, data, and models are in \url{https://github.com/FreedomIntelligence/AceGPT}.

\end{list}
\end{abstract}

\section{Introduction}

LLMs like GPT-3.5 Turbo and GPT-4 have been shaping the current landscape of natural language understanding and generation~\cite{bubeck2023sparks}. In contrast to the proprietary nature of GPT-3.5 Turbo and GPT-4, there has been a trend towards developing open-source large language models capable of instruction-following~\cite{alpaca} and fluent conversations~\cite{vicuna2023}, a phenomenon termed as `Democratization of ChatGPT’~\cite{DatabricksBlog2023DollyV2,touvron2023llama}. While these models have shown great promise in understanding and producing content in various languages, they might fail to align with local values and cultural norms in non-English environments~\cite {chen2023phoenix}; we call it the `localization issue'. This issue can lead to significant problems in practical usage scenarios, especially for regions such as the Arabic world where the culture and values diverge significantly from Western norms. 
We argue that it is not just desirable but necessary to localize large language models and tailor them to a specific cultural environment.

\paragraph{Methodology} The core of our approach lies in localizing large language models to the Arabic language using a packaged solution (known as \textbf{AceGPT}).
Firstly, through incremental pre-training on Arabic data (\textbf{localized pre-training}), we ensure that the model has a strong foundation in the Arabic language, including grammar, vocabulary, and cultural context. Next, by fine-tuning Arabic natural questions (\textbf{localized instructions}), we enable the model to effectively comprehend and respond to specific instructions that are pertinent to Arab interests. Furthermore, by generating Arabic native responses from GPT-4 (\textbf{localized responses}) rather than relying on translations from other languages, we ensure that the model's outputs are natural and fluent within an Arabic context thanks to the powerful GPT-4. Lastly, by employing a reward model based on \textbf{localized preference data} that respects local culture and value, we further refine the model to align the responses with the cultural and value norms of Arabic-speaking communities. 

\paragraph{Evaluation} 
We evaluate our models in various benchmarks:
in the  \textbf{instruction-following benchmark}, AceGPT achieves state-of-the-art (SOTA) among open-sourced Arabic LLMs in Arabic Vicuna-80 and Arabic AlpacaEval, obtaining 33\% and 30\% improvement over the state-of-the-art Arabic LLM~\cite{sengupta2023jais} \footnote{Jais \cite{sengupta2023jais} is a concurrent work released two weeks ahead of ours.}.
In the NLU benchmark, AceGPT achieves the second best on ALUE~\cite{seelawi-etal-2021-alue} in terms of average scores for all tasks.
In the \textbf{knowledge benchmark}, AceGPT achieves SOTA among open-sourced Arabic LLMs in Arabic knowledge including MMLU and EXAMs. 
In the \textbf{localization benchmark}, AceGPT achieves SOTA among open-source Arabic LLMs in our Arabic Cultural and Value Alignment~(ACVA) Dataset.

\paragraph{Contributions} The contributions of the paper are three-fold, including:
1) We propose a first-tier Arabic LLM. According to the records on the releasing date, it achieves SOTA performance among open Arabic LLMs in many benchmarks including Arabic Vicuna-80, Arabic AlpacaEval, Arabic MMLU, EXAMs, and ACVA.
2) AceGPT is the first open-source Arabic large language model that encompasses the entire LLM pipeline including pre-training, supervised fine-tuning, and reinforcement learning from AI feedback.
3) We observe and measure the localization issue in Arabic LLMs quantitatively, especially with a new benchmarking dataset, ACVA, for localization testing.

\section{Recipe of AceGPT}
\subsection{Motivation: The Localization Issue}

\begin{table}[t]
	\small
	\centering
	\caption{Proportion of Arabic Entities in Responses to 20 Sample Arabic Questions}
	\begin{threeparttable}
		\begin{tabular}{p{1.3cm}p{1cm}p{1.2cm}p{1cm}p{1cm}}
			\toprule
			Types of entity & Jais-13B & GPT-3.5 Turbo & GPT-4 & AceGPT (ours)\\
			\midrule
			Person &  \begin{tabular}[c]{@{}l@{}} 12.00\%\\  (3/25)\tnote{1} \end{tabular}{}{}    & \begin{tabular}[c]{@{}l@{}}  26.67\%\\  (12/45) \end{tabular}{}{}   & \begin{tabular}[c]{@{}l@{}}   39.29\%\\  (22/56) \end{tabular}{}{}  &\begin{tabular}[c]{@{}l@{}}  50.00\%\\  (31/62) \end{tabular}{}{}  \\
			\midrule
			Location & \begin{tabular}[c]{@{}l@{}}  18.75\%\\ (3/16) \end{tabular}{}{}   &  \begin{tabular}[c]{@{}l@{}}  27.08\%\\  (13/48) \end{tabular}{}{}   & \begin{tabular}[c]{@{}l@{}}  21.62\%\\   (16/74)\end{tabular}{}{} & \begin{tabular}[c]{@{}l@{}}  28.95\%\\   (11/38)\end{tabular}{}{}   \\
			\bottomrule
		\end{tabular}
		\begin{tablenotes}
			\item[1] \small 25 person names in Jais-13B responses are identified and 3 are Arabic names.
		\end{tablenotes}
	\end{threeparttable}
	\label{tab: Localization_Issue}
\end{table}

\begin{table*}[t]
	\small
	\caption{Comparison of LLM Responses Highlighting Location, Person, and Regional/Cultural Terms. Arabic-rooted terms in \textcolor{blue}{blue}; non-Arabic terms in \textcolor{red}{red}.}
	\label{tab: sample question response}
	\begin{tabularx}{\textwidth}{X|X}
		\toprule 
		\multicolumn{2}{p{13.4cm}}{
			\textbf{User}:
			\begin{RLtext}\tiny
				ما هي الكتب المقدسة والقديسين والأماكن المقدسة؟
			\end{RLtext}
			(What are the holy books, saints, and holy places?)} \\
		\midrule
		\textbf{Jais-13B chat}:
		\begin{RLtext}\tiny
			الكتاب المقدس هو مجموعة من النصوص الدينية التي تعتبر مقدسة لدى المسيحيين. القديسون هم الأشخاص المقدسون الذين يعتبرون أنهم عاشوا حياة فاضلة بشكل استثنائي ويتم تبجيلهم كقديسين من قبل بعض الأديان. الأماكن المقدسة هي مواقع تعتبر مقدسة أو مكرسة من قبل بعض الأديان ، مثل الكنائس والمعابد والمساجد
		\end{RLtext}
		( {\color{red} The Bible} is a collection of religious texts that are considered sacred by {\color{red}Christians}. Saints are holy people who are considered to have lived exceptionally virtuous lives and are venerated as saints by some religions. Holy places are sites considered sacred or consecrated by some religions, such as churches and temples And mosques.)
		&
		\textbf{GPT 3.5 Turbo}:
		\begin{RLtext}\tiny
			... في الديانات السماوية الثلاث (اليهودية والمسيحية والإسلام)، تعتبر الكتب المقدسة هي: اليهودية: التوراة (العهد القديم)،.... المسيحية: الكتاب المقدس المسيحي يتكون من العهد القديم (التوراة) والعهد الجديد ... الإسلام: القرآن الكريم، وهو الكتاب المقدس للمسلمين ويعتقد أنه وحي من الله للنبي محمد ... بعض الأماكن المقدسة الشهيرة تشمل: الكعبة في مكة المكرمة (الإسلام)، الحائط الغربي في القدس (اليهودية)، الكنيسة القديسة العائدة في بيت لحم (المسيحية)، والمسجد الأقصى في القدس (الإسلام).  
		\end{RLtext}
		(... In the three monotheistic religions (\textcolor{red}{Judaism}, \textcolor{red}{Christianity}, and \textcolor{blue}{Islam}), the holy books
		are: \textcolor{red}{Judaism}: \textcolor{red}{The Torah (Old Testament)} ... \textcolor{red}{Christianity}: The
		\textcolor{red}{Christian Bible} consists of the Old \textcolor{red}{Testament (Torah)} and the \textcolor{red}{New Testament}. ... \textcolor{blue}{Islam}: The \textcolor{blue}{Holy Qur’an}, which is the holy book of \textcolor{blue}{Muslims} and is
		believed to be a revelation from God to the \textcolor{blue}{Prophet Muhammad}.... the \textcolor{blue}{Kaaba} in \textcolor{blue}{Mecca} (Islam), the \textcolor{red}{Western Wall} in \textcolor{blue}{Jerusalem} (Judaism) ... \textcolor{red}{Bethlehem} (Christianity), and \textcolor{blue}{Al-Aqsa Mosque} in \textcolor{blue}{Jerusalem} (Islam).)\\
		\bottomrule
	\end{tabularx}
\end{table*}

Given the availability of many high-quality instruction datasets in widely spoken languages such as English, existing strategies for non-English LLMs often rely on instructions translated from English. Examples include Chinese-alpaca-GPT4~\cite{peng2023instruction}, Phoenix \cite{Chen_MultilingualSIFT_Multilingual_Supervised_2023}, and Jais~\cite{sengupta2023jais}.
However, relying on translated data may lead to \textit{localization issues}, potentially undermining the integrity and applicability of the models in native contexts.

To address these localization issues, we formulate 20 questions (see Table.\ref{20 sample questions}) to elicit responses with name entities—both personal and locational—to summarize the prevalence of Arabic name entities for preliminary experiments. Quantitative results in Table~\ref{tab: Localization_Issue} uncovers a significant deficiency in localization, where Jais-13B and GPT-3.5 Turbo only incorporate 12.00\% and 26.67\% Arabic names out of all the names in their responses respectively. A specific example is shown in Table~\ref{tab: sample question response}, we can observe that the Arabic open-source LLM Jais's output shows a conspicuous tilt towards English-centric materials, yielding terms predominantly associated with Christianity, which potentially neglects significant parallels within Arabic literary traditions. By contrast, GPT-3.5 Turbo showcases a more diverse recognition of holy sites from different cultural backgrounds. You can see the details and more examples of case studies in Appendix~\ref{subsec: case study}.

\subsection{Methodology of AceGPT}
To address localization, we propose a comprehensive solution including three strategies to ensure model's effective understanding and generation of content in Arabic, with cultural awareness and value alignment~\footnote{See the discussion on the relationship between Arabic culture and Arabic language in Appendix~\ref{sec:relationship}.}: \textbf{(I) localized pre-training} we further pre-train LLM with Arabic data; \textbf{(II) localized instructions} we adopt Arabic natural questions in the wild and their responses are Arabic native responses from GPT-4 instead of translating that from other languages, and \textbf{(III) localized responses} we further tame LLM with reinforcement learning using a reward model that respects local culture and values thanks to the \textbf{localized preference data}. 

The resultant model is termed `AceGPT'. The model pre-trained on LLaMA2~\cite{touvron2023llama} is named `AceGPT-base'. To equip it with the conversation, we introduced `AceGPT-chat' utilizing supervised fine-tuning and reinforcement learning from AI feedback. 
The training procedure is divided into three stages: pre-training, supervised fine-tuning, and reinforcement learning from AI feedback, introduced in Sec~\ref{sec_pretraining}, Sec~\ref{sec:sft}, and Sec~\ref{sec:rlhf}, respectively.

\subsubsection{Localized Pre-Training}
\label{sec_pretraining}

To adapt the English-focused LLaMA2~\cite{touvron2023llama} model in Arabic, we train further it with a substantial corpus of Arabic text.

\textbf{Data}
The dataset comprises Arabic and English sub-datasets. The Arabic is derived from the open-source Arabic text 2022~\footnote{\url{https://data.baai.ac.cn/details/ArabicText-2022} provided by BAAI}, and refined from sources like Arabic Wikipedia, CC100, and OSCAR3. The English dataset is obtained from Slim Pajama \cite{cerebras2023slimpajama} to avoid forgetting knowledge of English texts. Given LLaMA2's excellent adaptability to the English dataset, we sample a subset of data from Slim Pajama randomly. 

Due to the limit of computing resources, we only train the LLaMA2-7B with 30B data (19.2B tokens in Arabic and 10.8B in English) and LLaMA2-13B with 10B data (6B tokens in Arabic and 4B in English), prioritizing a larger quantity of Arabic data than English data. We utilized the original vocabulary of LLaMA2 which contains all 53 Arabic letters; The reason why we did not expand the vocabulary as existing work is to save training costs.

\subsubsection{Localized Supervised Fine-Tuning}
\label{sec:sft}

To enable the model to follow Arabic user instructions and tackle realistic applications, we fine-tuned AceGPT with localized instructions and localized responses. 

\textbf{Localized instructions and localized responses} The localized instructions are Arabic natural questions derived from real-world contexts, i.e. online question-answering platforms Quora~\footnote{\url{https://quora.com/}}, which can help models to capture what Arabs care in the wild. We can see from Table~\ref{tab:count_names_dataset} that common entities in the popular open-source datasets such as Alpaca are mostly Western (e.g. `John', `Apple', and `New York'), deviating from Arab's actual interest (e.g. `Mohammed', `Muslim Brotherhood', and `Egypt') which can be addressed by Quora. 
The main idea of localized responses is to leverage the fact that GPT-4 produces culture- and value-relevant responses in the context of question language, which means responses to questions in English are different from those in Arabic. See an example in Table~\ref{tab:culture_diff_example}, GPT-4 produces culture-dependent responses based on the queried languages. Therefore, when incorporating open-source instruction-tuning data, we ask the GPT-4 to re-generate responses in Arabic (rather than translate) to produce localized responses.

\begin{table*}[t]
	\small
	\centering
	\setlength{\tabcolsep}{2pt}
	\caption{ Top 5 names of individuals, organizations, and geopolitical entities (GPE) by frequency.}
	\label{tab:count_names_dataset}
		\begin{tabular}{l|l|l|l}
			\toprule 
			Dataset & Top-5 Person & Top-5 Organization & Top-5 GPE \\
			\midrule
			Alpaca & \makecell[l]{John, John Smith, Alice,\\Mary, Harry Potter} & \makecell[l]{Apple, Amazon, Google,\\Microsoft, ABC} & \makecell[l]{United States, India, New York,\\France, China} \\ 
			\midrule
			Evol-Instruct & \makecell[l]{John, John Smith, Harry Potter,\\Alice, Bob} & \makecell[l]{Apple, Amazon, quantum,\\Google, Microsoft} & \makecell[l]{United States, New York,\\Los Angeles, San Francisco,\\Japan} \\ 
			\midrule
			ShareGPT & \makecell[l]{Di Maria, Messi, Beckhaus,\\Eco, Clara} & \makecell[l]{Tribunal, Google, Council,\\Bing, Supreme Court} & \makecell[l]{United States, Argentina, France,\\New York, Hong Kong} \\ 
			\midrule
			Quora & \makecell[l]{Prophet, Mohammed, Adam,\\Hijri, Ali} & \makecell[l]{European Union, Google\\Muslim Brotherhood,\\Soviet Union, United Nations} & \makecell[l]{Egypt, Turkey, Saudi Arabia,\\Morocco, America} \\ 
			\bottomrule
		\end{tabular}
\end{table*}

\begin{table*}[t]
	\small
	\caption{GPT-4 answers culture-relevant questions differently across languages. Questions here are the same in semantics but differ in languages. The Arabic response is translated into English (right).}
	\label{tab:culture_diff_example}
	\begin{tabularx}{\textwidth}{X|X}
		\toprule 
		\textbf{Question in English}: & \textbf{Question in Arabic}:
		\\	
		Is it forbidden for a woman to confess her love to a man? &
		\begin{RLtext}\tiny
			هل من الحرام أن تعترف المرأة بحبها للرجل؟
		\end{RLtext}
		\\
		\textbf{GPT-4}:
		No, it is not forbidden for a woman to confess her love to a man. Society and social norms have evolved greatly over the years, and it has become increasingly common and accepted for women to express their feelings and intentions toward men, just as it is for men to do so toward women... &
		\textbf{GPT-4}:
		(In Islam, it is not necessarily considered haram for a woman to express her feelings towards a man, but it must be done decently and respectfully and in accordance with Islamic teachings. Islam encourages modesty and chastity, and therefore both men and women must display decency and good morals in their dealings. If a woman wants to express her feelings, she can do so directly or through an intermediary, such as her family...) \\
		\bottomrule
	\end{tabularx}
\end{table*}

\textbf{Data} In addition to Arabic Quora questions, we also incorporate some open-source instruction-tuning datasets to improve the overall performance. Specifically, we incorporate Alpaca~\cite{alpaca,peng2023instruction} (the most classical instruction-tuning dataset), Evol-Instruct~\cite{xu2023wizardlm} (a complex instruction dataset), Code-Alpaca~\cite{codealpaca} (a code-specific instruction dataset)~\footnote{We incorporate code-alpaca for a more powerful LLM with a better code capability.}, and ShareGPT~\footnote{\url{https://huggingface.co/datasets/philschmid/sharegpt-raw}} (a popular user-GPT dialogue dataset). For these open-source data except  ShareGPT, an Arabic version is created by translating the English questions into Arabic and re-generating the responses using GPT-4. We reserve the original ShareGPT data because the original conversations will be destroyed with a re-generated different response.

\subsubsection{Reinforcement Learning from   AI Feedback}
\label{sec:rlhf}

To further align AceGPT with values and cultures, we utilize reinforcement learning from AI feedback with a reward model trained with localized preference data. There are primarily two stages: (1) training the reward model using localized preference data, and (2) aligning AceGPT to value and culture preference patterns using the proximal policy optimization algorithm~\cite{Schulman2017}.

\textbf{Localized preference data}
To align AceGPT with Arabic culture and values, a reward model mimicking the preferences of native speakers is essential. To prepare the localized preference data for reward model training, we reuse 40K localized instructions, i.e. Quora questions, in the SFT stage and sample paired outputs from our fine-tuned 7B model.
Given the resource-intensive nature of collecting human feedback, we utilized GPT-4 feedback, which has been shown to correlate highly with human preference labeling and achieves competitive performance in text summarization~\cite{lee2023rlaif}. However, due to observed position bias in GPT-4~\cite{zhang2023huatuogpt}, we altered the order of sample answers and retained consistent preferences between two order-switched runs, resulting in 12K pairs. A small study with 800 examples verified the reliability of this preference data, revealing a correlation coefficient of 0.84 between GPT-4 and human evaluations. We also incorporate 12K open-source preference data for better generalization.   See Appendix~\ref{app:rlhf} for details.

\textbf{Reward model} The reward model operates within a `binary' framework, determining preferences with an additional linear head post the final hidden states. The loss function is expressed as:
\begin{equation}
	\small
	\mathcal{L}(\theta) = -\frac{1}{\|D\|}\mathbb{E}_{(x,y_c,y_r)\sim D}\left[ \log(\sigma(r_\theta(x,y_c)-r_\theta(x,y_r))) \right].
\end{equation}
Here, $x$ is the input, $y_c$ is the chosen model output, $y_r$ is the rejected model output of the pair, and $r_\theta$ is the reward model with the parameter $\theta$.

\textbf{Proximal policy optimization}
We crawl another 30K Quora questions different from Quora-40K for PPO training data. Proximal Policy Optimization (PPO) is an off-policy policy gradient method for reinforcement learning~\cite{Schulman2017}. The policy $\pi_\theta(a|s)$ represents the probability distribution over the next token $a$ given a sequence of previous tokens $s$, where $\theta$ are the model parameters. The primary objective is to maximize the preference signal from the reward model that corresponds to the desired output behaviour. The objective can be shown as
\begin{equation}
\small
	\mathcal{L}(\theta) = \mathbb{E}_{t}\left[ \min\left( \rho A_{t}, \text{clip} \left( \rho, 1-\epsilon, 1+\epsilon \right) A_{t} \right) \right], 
\end{equation}
where $\rho$ is a factor defined by $\rho=\pi_{\theta}(a_{t}|s_{t})/\pi_{\theta_{\text{old}}}(a_{t}|s_{t})$, $\theta_{\text{old}}$ denotes the model parameter used for experience sampling. $A_t$ refers to the advantage function that measures the relative value of generating $a_t$ as the next token conditioned on the sequence $s_1 \cdots s_t$, and $\epsilon$ is a hyperparameter for stability.

\section{Localization Evaluation}
\subsection{Evaluation Protocol}
In this section, we delve into the `Localized Evaluation' of our language model, focusing exclusively on the Arabic Cultural and Value Alignment (ACVA). ACVA serves as a critical benchmark to assess our model's performance in terms of its alignment with Arabic cultural nuances and values. This evaluation is particularly important for understanding how well the model adapts to the specific linguistic and cultural context of the Arabic language. We conduct this evaluation using our fine-tuned chat models, which have been specifically optimized for higher relevance and accuracy in culturally specific scenarios.

\textbf{Arabic Cultural and Value Alignment (ACVA)}
ACVA is a Yes-No question dataset, comprising over 8000 questions, generated by GPT-3.5 Turbo from 50 designed Arabic topics to assess model alignment with Arabic values and cultures (see Appendix~\ref{app: ACVA_construction} for data construction details and show some examples in Table \ref{AVCA eg}). A subset, revised by Arabic speakers for question quality and answer accuracy, forms the 2486-data `Clean set'. The correlation between `All set' and `Clean set' evaluations is in Sec~\ref{sec:acva_experiment_results}. Given our focus on localized solutions, we evaluate our final models (post-SFT and RLAIF) on this benchmark in a zero-shot setting with F1 score.

\textbf{Baselines}
We compare the performance of our models against LLaMA2~\cite{touvron2023llama}, Bloomz~\cite{muennighoff2022crosslingual}, Phoenix~\cite{chen2023phoenix, Chen_MultilingualSIFT_Multilingual_Supervised_2023}, and Jais~\cite{sengupta2023jais} for this section and Sec~\ref{sec: evaluation}.
LLaMA2-chat models are excluded as they consistently respond in English when queried in Arabic. See details in Appendix~\ref{sec: baselines}.
\subsection{Experiment Results}
\label{sec:acva_experiment_results}
\begin{table}[t]
	\small
	\caption{Average F1 on ACVA in the zero-shot setting. The best performance is in \textbf{bold} and the second is \underline{underlined}.}
	\label{tab: local-details}
	\centering
	\begin{tabular}{lll}
		\toprule
            {Model} &All Set & Clean Set \\ 
		\midrule
		Phoenix   & 41.86\% &43.80\% \\
		Phoenix--multiple-langs & 59.78\% &59.15\% \\
		Jais-13B-chat   & 61.44\% & 66.83\%  \\ 
		\midrule
		\textbf{AceGPT-7B-chat}     & 69.53\% & 70.03\%\\
		\textbf{AceGPT-13B-chat}      & \underline{75.02\%} &\underline{74.62\%} \\ 
		\midrule
		GPT-3.5 Turbo          &\textbf{75.57\%} & \textbf{79.03\%} \\
		\bottomrule   
	\end{tabular}
\end{table}
\textbf{ACVA benchmark}
We present the results of AceGPT and other chat models on ACVA in Table~\ref{tab: local-details}. The Pearson correlation of accuracy on `All set' and `Clean set' is $0.9825$, indicating a high reliability of ACVA all set evaluation. Notably, our AceGPT-chat models (both 7B and 13B) consistently outperform other open-source LLMs, and AceGPT-13B-chat only trails GPT-3.5 Turbo by a marginal of $-0.55\%$ on all set. Since the Jais-30B-chat does not follow instructions and cannot return answers for multiple-choice questions, we suspect that this is due to overly stringent safety measures. Therefore, we did not include Jais-30B-chat in the zero-shot comparison.

\section{Overall Evaluation}

\begin{table}[t]
    \small
    \caption{Average performance ratio of GPT-3.5 Turbo and the standard variation over three runs in Arabic Vicuna-80 and Arabic AlpacaEval. The best performance is in \textbf{bold} and the second is \underline{underlined}.}
    \label{tab:vicuna80_gpt4}
    \centering
    \setlength{\tabcolsep}{3pt}
    \resizebox{\columnwidth}{!}{
    \begin{tabular}{lrr}
        \toprule
        Comparison & \begin{tabular}[c]{@{}c@{}}Arabic Vicuna-80\end{tabular} & \begin{tabular}[c]{@{}c@{}}Arabic AlpacaEval\end{tabular} \\
        \midrule
        Phoenix & 71.92\% ± 0.2 & 65.62\% ± 0.3 \\
        Phoenix (multi-langs) & 71.67\% ± 0.7 & 65.36\% ± 0.1 \\
        Jais-13B-chat & 75.40\% ± 1.6 & 74.95\% ± 0.2 \\
        \midrule
        \textbf{AceGPT-7B-chat} & \underline{94.82}\% ± 0.2 & \underline{93.81}\% ± 0.1 \\
        \textbf{AceGPT-13B-chat} & \textbf{100.88}\% ± 0.4 & \textbf{97.95}\% ± 0.1 \\
        \bottomrule
    \end{tabular}
    }
\end{table}

\subsection{Evaluation Protocol}
\label{sec: evaluation}

Evaluation of language models is multifaceted and typically involves multiple metrics and benchmarks to assess various aspects of model performance. Moving beyond the scope of localization, the `Overall evaluation' section presents a comprehensive analysis of our language model across a spectrum of benchmarks. This includes assessing instruction-following ability, knowledge retention, and Natural Language Understanding (NLU). For evaluating instruction-following ability, we employ our fine-tuned chat models, which are designed to excel in interactive and directive tasks. In contrast, knowledge retention and NLU are evaluated using our base models, focusing on the core strengths of the model's pre-training. While we utilize both automated and manual methods for assessing instruction-following ability, other benchmarks in this section are evaluated solely through automated methods.

\textbf{Instruction-following capability}
We specifically assess the models' instruction-following capabilities using the Arabic versions of Vicuna-80~\cite{vicuna2023} and AlpacaEval~\cite{dubois2023alpacafarm}, translated by GPT-4 and refined by native speakers.  In accordance with~\cite{vicuna2023}, we adopt the GPT-4 evaluation, which prompts GPT-4 to score the performance of models on each question, contrasting them with GPT-3.5 Turbo.
The details can be found in Appendix~\ref{app: Evaluating prompt gpt4}. While GPT-4 evaluation is efficient and scalable, it may overlook the subtle inconsistencies between model responses~\cite{wang2023evalfair} and human interactions in real-world scenarios. Therefore, we further conduct human evaluation on these benchmarks to evaluate the performance of AceGPT from the perspective of human rather than GPT-4 preferences. 
To ensure cultural relevance in manual evaluations, we engaged a diverse group of educated, native Arabic speakers. Each model's response was assessed independently by three assessors. We present more details in Table~\ref{app: background} and the UI for evaluation in Figure~\ref{fig: ui}.

\textbf{Knowledge benchmark} We have two knowledge benchmarks, including Arabic MMLU and EXAMs.

MMLU~\cite{MMLU21} consists of diverse multiple-choice questions across 57 tasks, spanning various educational levels. We employed GPT-3.5 Turbo to translate this dataset from English to Arabic. Additionally, Arabic questions from the EXAMs~\cite{EXAMS20}, a resource specialized in multilingual high school exam questions, were also incorporated. Both datasets were evaluated in a few-shot setting, as per the methodology in~\cite{huang2023ceval}, to assess the innate capabilities of LLMs, aiming at potential applications with minimal adaptations.

\textbf{ALUE}~\footnote{\url{https://www.alue.org/home}\label{web-alue}} is a popular online benchmark, which is similar to the GLUE benchmark but has a main focus on \textbf{A}rabic \textbf{L}anguage \textbf{U}nderstanding \textbf{E}valuation. It includes traditional NLP tasks such as sentiment analysis, semantic matching, text relation classification, and dialect identification. It comprises 9 tasks as illustrated in Table~\ref{table:nlp-tasks}. More details of task illustration, experiment setting, and results can be found in Appendix~\ref{sec: NLU-evalution}.

\subsection{Experiment Results}
\label{sec:experiment_results}

\textbf{Instruction-following benchmark}
We present each model's performance ratio against GPT-3.5 Turbo, scored by GPT-4, in Table~\ref{tab:vicuna80_gpt4}. The result shows that AceGPTs are superior in both Arabic Vicuna-80 and Arabic AlpacaEval. Notably, AceGPT-7B-chat surpasses Jais-13B by about 20\% points with smaller model size. Moreover, AceGPT-13B-chat attains a 100.88\% performance ratio of GPT-3.5 Turbo in Arabic Vicuna-80.

\textbf{Human evaluation}
Table~\ref{tab:human_eval_result} shows the human evaluation results on Arabic Vicuna-80 and Arabic AlpacaEval. We calculated the percentages of wins, ties, and losses of the results from three Arabic speakers. We observe that AceGPT-chat (both 7B and 13B) significantly surpasses Jais-13B-chat and even Jais-30B-chat, but lags behind GPT-3.5 Turbo. Moreover, the AceGPT-13B-chat is significantly better than the AceGPT-7B-chat, indicating the importance of model size.

\begin{table*}[t]
	\small
	\caption{Human evaluations on Vicuna-80 and AlpacaEval. The winners are in \textbf{bold}.}
	\label{tab:human_eval_result}
	\centering
	\begin{tabular}{l|llll|c}
		\toprule
		Dataset & Comparison & Win & Tie & Lose& Win or Tie\\ 
		\midrule
		\multirow{6}{*}{Arabic Vicuna-80}&\textbf{AceGPT-7B-chat} vs. Jais-13B-chat & 82.5\%& 6.7\% &10.8\%&89.2\%\\
		&AceGPT-7B-chat vs. \textbf{GPT-3.5 Turbo}& 27.5\% &32.9\% &39.6\%&60.4\%\\
		\cmidrule(l){2-6}
		&\textbf{AceGPT-13B-chat} vs. Jais-13B-chat & 82.9\%& 6.7\% &10.4\%&89.6\%\\
		&AceGPT-13B-chat vs. \textbf{GPT-3.5 Turbo} & 16.3\% &57.1\% &26.6\%&73.4\%\\
            \cmidrule(l){2-6}
            &\textbf{AceGPT-7B-chat} vs. Jais-30B-chat & 67.5\%& 15.0\% &17.5\%&82.5\%\\
		&\textbf{AceGPT-13B-chat} vs. Jais-30B-chat& 64.6\% &15.0\% &20.4\%&79.6\%\\
		\midrule
		\multirow{4}{*}{Arabic AlpacaEval}&\textbf{AceGPT-7B-chat} vs. Jais-13B-chat & 53.0\%& 36.5\% &10.5\%&89.5\%\\
		&AceGPT-7B-chat vs. \textbf{GPT-3.5 Turbo}& 20.2\% &46.5\% &33.3\%&66.7\%\\
		\cmidrule(l){2-6}
		&\textbf{AceGPT-13B-chat} vs. Jais-13B-chat & 49.4\%& 42.8\% &7.8\%&92.2\%\\
		&AceGPT-13B-chat vs. \textbf{GPT-3.5 Turbo} & 25.2\% &44.5\% &30.3\%&69.7\%\\
		\bottomrule   
	\end{tabular}
\end{table*}

\textbf{Knowledge benchmark}
Table~\ref{tab:mmlu_label} shows the few-shot evaluation results on Arabic MMLU and EXAMs. We can see that AceGPT-13B-base attains the best performance in Arabic MMLU (37.26\%) among open-source LLMs, and AceGPT-7B-base also surpasses other open-source models, including 13B models, in Humanities and Others (Business, Health, Misc) domains in Arabic MMLU. 
In EXAMs, the Jias-30B-base model achieves the best performance among open-source models.

\begin{table*}[t]
	\small
	\caption{Accuracy on Arabic MMLU and EXAMs. The best is \textbf{bold} and the second is \underline{underlined}. }
	
	\label{tab:mmlu_label}
	\centering
	\begin{tabular}{l|ccccc|c}
		\toprule
		\multirow{2}{*}{Model} & \multicolumn{5}{c}{Arabic MMLU} & \multirow{1}{*}{EXAMs} \\
		& Average & STEM & Humanities &Social Sciences  & Others \\ 
		
		\midrule		
		Bloomz & 33.69 & 33.35 & 29.29 & 37.58 & 34.53 & 33.89 \\
		LLaMA2-7B & 29.47 & 30.30 & 29.33 & 27.46 & 30.78 & 23.48 \\
		LLaMA2-13B & 33.76 & 32.94 & 32.30 & 33.42 & 37.27 & 25.45 \\
		Jais-13B-base & 32.23 & 30.51 & 31.25 & 33.74 & 33.43 & 35.67 \\
            Jais-30B-base & 36.27 & 32.67 & 30.67 & 42.13 & 39.60 & \underline{39.91} \\
		\midrule
		AceGPT-7B-base & 32.14 & 29.73 & 30.95 & 33.45 & 34.42 & 31.96 \\
		AceGPT-13B-base & \underline{40.45} & \underline{36.60} & \underline{38.74} & \underline{43.76} & \underline{42.72} & 36.63 \\ 
		\midrule
		GPT-3.5 Turbo & \textbf{49.07} & \textbf{43.38} &\textbf{ 44.12} & \textbf{55.57} & \textbf{53.21} & \textbf{45.63} \\
		\bottomrule   
	\end{tabular}
	
\end{table*}

\section{Experimental Analysis}

\subsection{On  Pre-Training}

\textbf{Localization of pre-training}
AceGPT-base uses LLaMA2 as the backbone, the only difference it is further pre-trained with some local Arabic texts.
We compare AceGPT-base to LLaMA2 on ACVA with the few-shot setting to demonstrate the benefits of localized pre-training on Arabic culture and values. The results in Table~\ref{tab:ablation_pretraining} show the superiority of localized pre-training: after localized pre-training, AceGPT-7B-base surpasses LLaMA2-13B, which has a larger size.
\begin{table}[h]
	\small
	\caption{Ablation of   pre-training. 
	}
	\label{tab:ablation_pretraining}
	\centering
 \vspace{-8pt}
	\begin{tabular}{llc}
		\toprule
		Size & Model     & F1  on ACVA \\
		\midrule
		\multirow{2}{*}{7B}  & LLaMA2        & 51.44\% \\
		& AceGPT-base   & \underline{68.28}\% \\
		\midrule
		\multirow{2}{*}{13B}  & LLaMA2       & 65.67\% \\
		& AceGPT-base  & \textbf{76.23}\% \\
		\bottomrule
	\end{tabular}
\end{table}

\subsection{On Supervised Fine-Tuning}

\begin{table}[t] 
\small
	\caption{Effects of different datasets on ACVA, Arabic Vicuna-80 and Arabic AlpacaEval.}
	\label{tab:ablation_sft}
	\centering
 \resizebox{\columnwidth}{!}{
	\begin{tabular}{lllll}
		\toprule
		Comparison & ACVA & \begin{tabular}[c]{@{}l@{}}Arabic\\  Vicuna-80\end{tabular}{}{}  & \begin{tabular}[c]{@{}l@{}}Arabic\\  AlpacaEval\end{tabular}{}{}   \\ 
		\midrule
  Alpaca-Arabic & 50.52\% & 87.15\% ± 0.5 & 82.97\% ± 0.4  \\
  \midrule
		+ ShareGPT    & 38.64\% & 88.01\% ± 0.03 & 84.89\% ± 0.3    \\  
		\midrule
		+ Evol-Instruct & 61.72\% & \textbf{90.39\%} ± 0.4 & \textbf{86.87\%} ± 0.1   \\    
		\midrule
		+ Quora & \textbf{65.53\%}  & 89.74\% ± 0.8 & 85.71\% ± 0.03   \\
		\bottomrule   
	\end{tabular}
 }
\end{table}

 \begin{table*} 
	\small
	\caption{Experiments with/without RLAIF on Arabic Vicuna-80, Arabic AlpacaEval and ACVA.}
	\label{tab: ablation}
	\centering
	\setlength{\tabcolsep}{2pt}
	\begin{tabular}{l|lll|llcccc}
		\toprule
		& \multicolumn{3}{c}{Automatic Evaluation} & \multicolumn{4}{c}{Human Evaluation (vs. GPT-3.5 Turbo)} \\ \midrule
		Comparison& ACVA  & Arabic Vicuna-80 & Arabic AlpacaEval & Win & Tie & Loss & Win or Tie \\ 
		\midrule
		AceGPT-7B-chat (w/o RLAIF)  & 42.48\%  & 92.01\% ± 1.3\%    & 91.35\% ± 0.08\% &27.5\% & 29.2\% & 43.3\% & 56.7\% \\
		AceGPT-7B-chat              & \textbf{69.53\%} & \textbf{94.82\%} ± 0.2\% & \textbf{93.81\%} ± 0.1\%  &27.5\% & 32.9\% & 39.6\% & 60.4\%\\
		\midrule
		AceGPT-13B-chat (w/o RLAIF) & 74.18\%  & 95.14\% ± 1.0\%    & 93.05\% ± 0.2\%  &19.6\% & 37.5\% &42.9\% & 57.1\% \\
		AceGPT-13B-chat             &\textbf{75.02\%} & \textbf{100.88\%} ± 0.4\% & \textbf{97.95\%} ± 0.1\%  &16.3\% & 57.1\% &26.7\% & 73.3\%  \\
		\bottomrule   
	\end{tabular}
\end{table*}
In this analysis, we primarily assess the impact of both localized and open-source instructions on localization and overall performance. Each dataset has been sampled with 40,000 data points, respectively.
The results are shown in Table~\ref{tab:ablation_sft}.
It can be observed that Evol-Instruct highly contributes to the overall performance in the instruction-following benchmark, while Quora is most beneficial for Arabic culture and values. Note that incorporating ShareGPT largely harms the performance of ACVA; this may be because  ShareGPT is almost aligned with Western culture and values.

\subsection{On RLAIF}

\subsubsection{Reward Model}

\begin{figure}[t] 
	\centering 
	\begin{subfigure}{.35\textwidth}
		\centering
		\includegraphics[width=.99\linewidth]{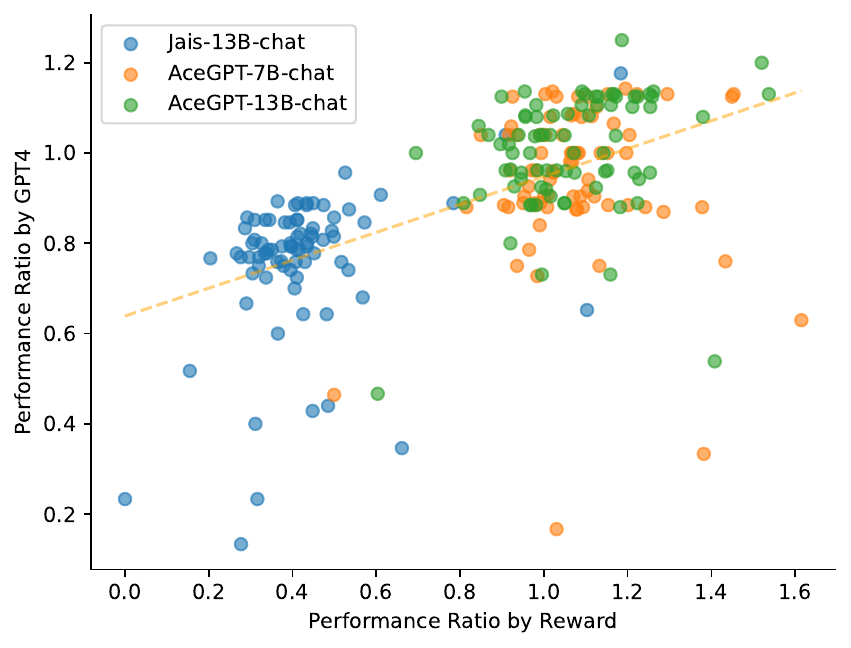} 
		\caption{Correlations between reward model scoring and GPT-4 scoring on Arabic Vicuna-80.}
		\label{fig:reward_correlations_vicuna}
	\end{subfigure}
	\hfill
	\begin{subfigure}{.35\textwidth}
		\centering
		\includegraphics[width=.99\linewidth]{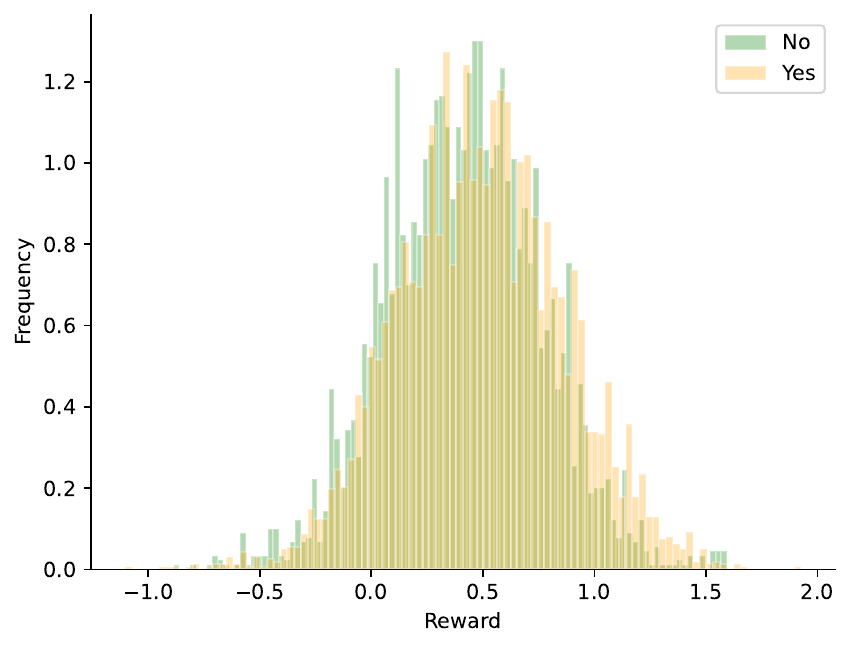}
		\caption{Reward score distribution on ACVA.}
		\label{fig:culture_reward_distribution}
	\end{subfigure}
	\caption{ (a) Correlations between the reward model and GPT-4 and (b) reward distribution.}
	\label{fig:combined_figure}
\end{figure}

To evaluate the sensitivity of the reward model to the overall performance, we measure the correlations between reward scoring and GPT-4 scoring (described in section~\ref{sec: evaluation}) on Arabic Vicuna-80. Following the pairwise comparison setting in GPT-4 scoring, we also calculate the performance ratio for normalized (to [0, 10] as GPT-4 scoring) reward scores on model-chatbot pairs. The Pearson correlation and Spearman correlation are 0.57 and 0.61 respectively, and the results are shown in Figure~\ref{fig:reward_correlations_vicuna}. We conclude that the reward model shows a positive correlation with GPT-4 evaluation on Arabic Vicuna, which indicates it can offer an effective signal on overall performance.

\begin{table}[t] 
\small
	\caption{Ratio of the frequency of "yes" to "no" of Reward.}
	\label{tab:Ratio of reward}
	\centering
 \resizebox{\columnwidth}{!}{
	\begin{tabular}{lllll}
		\toprule
		Reward interval	 & (-$\infty$,0) & (0, 0.5) &	(0.5, 1)	& (1,$\infty$) \\  
            Ratio of Yes/No	& 81.7\% & 86.4\%	&110.8\% &208.2\%	\\
		\bottomrule   
	\end{tabular}
 }
\end{table}

\textbf{Localization of reward model}
Then we evaluate the Arabic culture sensitivity of the reward model on the ACVA benchmark. Prompting with `Give me a fact about Arab culture, values, and laws' in Arabic, we calculate the reward scores of prompt-statement pairs for all statements from ACVA. The distribution of reward scores for yes/no statements is shown in Figure~\ref{fig:culture_reward_distribution}, and the ratio of the frequency of "yes" to "no" is shown in Table~\ref{tab:Ratio of reward}. It demonstrates that reward scores for `yes' statements are higher than `no' statements overall, which suggests that our reward model has a cultural sensitivity.

\subsubsection{Ablation}

To empirically validate the contribution of RLAIF on overall performance and localization to our AceGPT models, we conduct ablation studies across ACVA benchmarks, Arabic Vicuna-80 and Arabic AlpacaEval, results are outlined in Table~\ref{tab: ablation}.

\textbf{RLAIF improves localization}
RLAIF results in performance gains of 27.05\% and 0.84\% for AceGPT-7B and AceGPT-13B in ACVA respectively, despite not being explicitly trained for them. This suggests that RLAIF enhances alignment with Arabic culture and values. Notably, the improvement from RLAIF on the 7B model is much larger than that of 13B, partially because the 7b model is weaker and therefore has more space for improvement, while it may be in saturation in the 13B model. 
Another reason could be that the preference data in RLAIF are generated from AceGPT-7b and therefore the learned reward model fits better AceGPT-7b than AceGPT-13b.

\textbf{RLAIF improves instruction-following}
The results show that RLAIF significantly enhances overall model performance on both Arabic Vicuna-80 and Arabic AlpacaEval, increasing AceGPT-7B's performance by 2.81\% and 2.46\%, while AceGPT-13B shows an improvement of 5.74\% and 4.90\%, respectively.
By examining the `win or tie' metric,  the 7B model shows an enhancement of 3.7\% through RLAIF, while the 13B model shows a significant boost of 16.2\%. This narrows the gap with GPT-3.5 Turbo. These enhancements across datasets underscore RLAIF's efficacy.

\section{Conclusion}

AceGPT addresses the `localization issue' in large language models by specifically catering to the distinct linguistic and cultural contexts of Arabic environments, leveraging further pre-training, instruction tuning, and RLAIF. It excels in multiple benchmarks including instruction-following and natural language understanding, setting a new standard among Arabic LLMs. We contribute high-quality datasets and evaluation resources, highlighting the need for localizing large language models and introducing AceGPT as a pioneering solution for Arabic linguistic and cultural adaptation.

\section*{Limitation}

Our AceGPT model exhibits several limitations. Its vocabulary, is primarily focused on Arabic letters, lacking further expansion, affecting Arabic text encoding efficiency. Limited machine resources during pre-training restricted token allocation, suggesting untapped potential in Arabic content processing. We omitted reasoning/misinformation and bias testing in our evaluation, raising concerns about the model's safety alignment and currently limiting its use to academic research rather than online deployment. Additionally, despite manual checks, the cultural dataset requires enhancement in both quality and quantity, which may affect the model's practicality and adoption.

\section*{Ethical Statement}

We mainly use public data to train our models. For the newly-collected data (e.g., Quora), we use ChatGPT to filter the questions with any ethical issues.  Additionally, we have adopted RL to aligned with human values.

\section*{Acknowledgement}
This work was supported by Hetao Shenzhen-Hong Kong Science and Technology Innovation Cooperation Zone Project (No.HZQSWS-KCCYB-2024016), the Internal Project Fund from Shenzhen Research Institute of Big Data under Grant J00220230010,  the Shenzhen Science and Technology Program (JCYJ20220818103001002), Shenzhen Doctoral Startup Funding (RCBS20221008093330065), Tianyuan Fund for Mathematics of National Natural Science Foundation of China (NSFC) (12326608), and KAUST Baseline Research Fund.

We thank Prof. Zhi-Quan Luo and Dr. Ping Lee for their support. We extend our sincere appreciation to the dedicated KAUST graduate students whose contributions were integral to the success of our Arabic evaluations, including Lamees Alzahrani, Abdullah Amr Bawazir, Nouf Khalil Alenizi, Shatha Abdullah Alowdah, Rudaynah Maimani, Feras Khalid Alwutayd, Abdulrahman, Arwa Fallatah, Noura Alhijri, Reem Alquwayzani, and Majid Almarhoumi. We thank them for their invaluable support in this research.

\FloatBarrier

\bibliography{anthology,custom}

\clearpage
\newpage
\appendix

\section{Related Work}
Our work mainly lies in two topics: democratization of ChatGPT to low-resource language and localization and cultural alignment.

\paragraph{Democratization of ChatGPT to low-resource language} 
Recent trends have focused on developing open-source LLMs as an alternative to the proprietary GPT-3.5 Turbo and GPT-4 models (\citealp{alpaca,vicuna2023,DatabricksBlog2023DollyV2,chen2023phoenix,sengupta2023jais}). These efforts have expanded to include LLMs designed for languages beyond English (\citealp{chen2023phoenix,sengupta2023jais}). However, these models rely heavily on large-scale datasets translated from English in the supervised fine-tuning stage, inadvertently embedding English-centric cultural biases and values (\citealp{chen2023phoenix,sengupta2023jais}). This practice risks undermining the models' capacity to align with the cultural contexts and values specific to their target linguistic communities. To mitigate this ``localization issue'', our work proposes a culture-aware framework that encompasses the pretraining, supervised fine-tuning, and reinforcement learning phases, to enhance the value and cultural alignment of AceGPT with Arabic.

\paragraph{Localization and cultural alignment} 
Cultural diversity varies significantly by region and language. Yet, pre-trained language models primarily incorporate Western cultural biases (\citealp{PaltaR23fork,Ramezani023,CH-WangSLYM23,naous23}). These biases are attributed to the reliance on training corpora such as C4 (\citealp{DodgeSMAIGM021}) and Wikipedia (\citealp{CallahanH11}), which are inherently skewed towards Western culture, leading to English-centric models. Moreover, the emphasis on translating English data in low-resource language NLP research exacerbates the ``localization issue", failing to adequately represent the non-English culture and values (\citealp{conneau2018xnli, chen2023phoenix, sengupta2023jais}). To address this, it is recommended to develop culture-aware models for non-English languages by gathering and annotating culture-specific data (\citealp{HershcovichFLLA22}). However, assembling comprehensive datasets for low-resource languages is challenging and expensive. Previous research has either concentrated on collecting unsupervised data for pre-training (\citealp{sengupta2023jais}) or embarked on task-specific supervised fine-tuning (\citealp{wang23, conneau2018xnli, HuangY23, PaltaR23fork}). In this paper, we present a novel and comprehensive culture-aware pipeline that goes beyond the previous approaches, by integrating general culture-aware supervised fine-tuning and culture-aware reinforcement learning.

\section{Localization Issues}
\label{app: localization}

\subsection{Sample Questions for Localization}
The sample questions for Arabic name entity comparison in Table~\ref{tab: Localization_Issue} and \ref{tab: sample question response} are as follows Table~\ref{20 sample questions}.

\begin{table}[ht]
	\small
	\centering
	\setlength{\tabcolsep}{3pt}
		\caption{20 sample questions}
		\label{20 sample questions}
            \begin{tabular}{l}
		\toprule
		\tiny \<ما هي الأسماء العشرة الأكثر شيوعاً؟> \\
		What are the ten most common names? \\
		\tiny \<ما هي القصص التاريخية العشرة الأكثر شهرة؟> \\
		What are the ten most famous historical stories? \\
		\tiny \<أسماء المدن العشرة الأكثر شيوعاً> \\
		Names of the ten most common cities\\
		\tiny \<ما هي أكثر عشر مدن يسمع عنها الناس أكثر؟> \\
		What are the top ten cities people hear about the most? \\
		\tiny \<ما هي أهم عشر منظمات اجتماعية عربية؟> \\
		What are the ten most important Arab social organizations? \\
		\tiny \<أفضل عشرة أشخاص لهذا العام> \\
		Top Ten People of the Year\\
		\tiny \<كيف يحتفل الناس بالعام الجديد؟> \\
		How do people celebrate the New Year? \\
		\tiny \<لماذا يؤمن الناس بالدين؟> \\
		Why do people believe in religion?\\
		\tiny \<من هم العلماء المشهورون الجدد في الوطن العربي؟> \\
		Who are the new famous scholars in the Arab world? \\
		\tiny \<ما هي بعض المعالم السياحية الشهيرة؟ ما هي الأحداث المحلية؟> \\
		What are some famous tourist attractions? What are local\\ events?\\
		\tiny \<من هم الكتاب والشعراء المشهورين في التاريخ؟> \\
		Who are the famous writers and poets in history? \\
		\tiny \<ما هي بعض الأطباق الشهية التي لا يتم الاستمتاع بها إلا خلال المهرجانات الكبرى؟> \\
		What are some of the delicacies that are only enjoyed during\\ major festivals? \\
		\tiny \<ما هو تقاليدنا الأدبية؟> \\
		What is our literary tradition? \\
		\tiny \<ما هي العادات والأنشطة الأساسية المهمة في حفل الزفاف؟> \\
		What are the basic customs and activities that are important\\ at a wedding? \\
		\tiny \<ما هي الكتب المقدسة والقديسين والأماكن المقدسة؟> \\
		What are the Holy Books, Saints, and Holy Places?\\
		\tiny \<ما هي بعض الحكايات الشعبية؟> \\
		What are some folk tales? \\
		\tiny \<كيف نشأت لغتنا؟> \\
		How did our language originate? \\
		\tiny \<ما هي أهم المهرجانات؟ لماذا توجد هذه المهرجانات؟> \\
		What are the most important festivals? Why do these\\ festivals exist?\\
		\tiny \<من هم الأشخاص الذين يجب أن نتذكرهم؟> \\
		Who are the people we should remember? \\
		\tiny \<من هم بعض الفنانين المتميزين؟> \\
		Who are some of the distinguished artists?\\
		\bottomrule
		\end{tabular}
\end{table}

\subsection{Case Study}
\label{subsec: case study}

In this subsection, we analyze the performance of AceGPT by conducting a comparative analysis of its localization ability via case studies on the sampled 20 localization questions. Illustrated in Table~\ref{tab: additional examples}, we observed a larger proportion of Arabic events in AceGPT. 
The first example in Table~\ref{tab: additional examples} aligns with the instance illustrated in Table~\ref{tab: sample question response}. Both AceGPT and GPT-3.5 Turbo exhibit superior responses to the given query, significantly surpassing the answer provided by Jais. Specifically, AceGPT’s understanding of a `holy book' is not solely confined to the Bible; it demonstrates a nuanced acknowledgment that different regions, especially Arabic, have their respective sacred texts, reflecting a broad and inclusive comprehension of diverse religious traditions. This illustrates the advanced capability of AceGPT, akin to GPT-3.5 Turbo, in response generation for Arabic-speaking areas.

The second example exemplifies the capability of AceGPT to incorporate more Arabic elements when responding to historical questions. Specifically, AceGPT allocates a significant proportion of its responses, 4 out of 10, to Arabic historical figures. In contrast, GPT-3.5 Turbo only attributes 1 out of 10 responses to Arabs, while Jais exclusively presents choices associated with Western figures. This demonstrates that AceGPT has an inclination towards Arabic culture, emphasizing its capability to offer more Arabic culture-relevant responses in an Arabic context.

\section{Construction of ACVA}
\label{app: ACVA_construction}

We employ a top-down approach for the construction of the Arabic Cultural and Value Alignment benchmark. First, we gathered over 50 topic keywords (see Table~\ref{tab: ACVA topics}) representing various aspects of Arabic culture, including humanity, art, science, geography, history, manners, religion, and the influence between civilizations, sourced from several books on Arabic culture and values. Then, we query GPT-3.5 Turbo to generate 8000 data based on the given topic using the prompt shown below, where \texttt{topic} is the placeholder for the topic.

We further sample 50\% topics to verify the relevance of questions to Arabic cultures and values and the accuracy of the Yes-No labels, which were reviewed by Arabic speakers, leading to a `Clean set'.
\vspace{8mm}
\hrule
\begin{quote}
	I am collecting some supervised fine-tuning (sft) data about Arabic culture. It is about the knowledge of Arabic culture and manners. The data is some questions in the Arabic language with an id in the form of \{'id': '1','label':'xx' 'query':'xx'\}. I will give you a topic in Arabic culture. The 'id' is the index of the data. 'label' is the topic I give you. 'query' is some question statement about Arabic culture under that topic. The Data should be of no repetition with a balanced proportion of true and false. Now please generate 200 sft data in json in Arabic with the format under the topic of \texttt{topic}
	
\end{quote}
\hrule

\begin{table*}[ht]
	\small
	\centering
		\caption{Topics for ACVA construction}
		\label{tab: ACVA topics}
  \begin{tabular}{p{3cm}|p{12cm}}
		\toprule
		Country&Algeria, Bahrain, Comoros, Egypt modern, Iraq, Jordan, Kuwait, Lebanon, Libya, Mauritania, Morocco, Oman, Palestine, Qatar, Saudi Arabia, Somalia, Sudan, Syria, Tunisia, United Arab Emirates, Yemen\\
		\midrule
		Civilization Relation&Influence from ancient Egypt, Influence from Byzantium, Influence from China, Influence from Greece, Influence from Persia, Influence from Rome, Mesopotamia civilization\\
		\midrule
		Science and Humanity&Arabic Astronomy, Arabic Math, Arabic Medicine, Arabic Physics and Chemistry, Arabic Literature, Arabic Music, Arabic Philosophy, Arab Empire, Arabic Architecture, Arabic Art, Arabic Calligraphy, Arabic Geography, Arabic History, Arabic Language Origin \\
		\midrule
		Manners and Religion& Arabic Ceremony, Arabic Clothing, Arabic Culture, Arabic Food, Arabic Funeral, Arabic Ornament, Arabic Wedding, Mindset, Special Expression, Daily Life, Influence from Islam, Islam Branches and Schools, Islam Education, Islamic Law System\\
		\bottomrule
		\end{tabular}
\end{table*}

\section{Preference Data for RLAIF}
\label{app:rlhf}

The data comprises two parts: Arabic preference data and open-source English preference data. Outputs for Arabic preference data are sampled from our fine-tuned 7B model with a temperature of 1. The open-source English preference data is incorporated to improve the generalization capability of the reward model and alleviate GPT4-preference hacking. We randomly sample 12K from three public human-annotated datasets - Anthropic helpfulness and harmlessness~\cite{HH2022}, OpenAI Summarize~\cite{summarize2020}, and OpenAssistant Conversations (OASST1)~\cite{openassistant23}. 

The core idea of preference labeling for Arabic preference data is to use a GPT-4 model with prompts as an automatic annotator to assess two responses generated by the same model for a given question. However, a significant challenge emerges as GPT-4 often shows a marked preference for the first response, around 80\% of the time, with the exact percentage varying based on the specific prompt design. To counter this, we utilize an order-switch mechanism to ensure consistent preference data across two separate runs of GPT-4. In one run, two responses are placed arbitrarily, and in the other, their orders are switched. The prompt for labeling is shown below. \texttt{instruction}, \texttt{response 1}, and \texttt{response 2} are the placeholders for the input instructions and the two generated responses.

\vspace{25mm}
\hrule
\begin{quote}
	[Instruction] \\
	\texttt{instruction} \\

	[Response1] \\
	\texttt{response 1} \\
	
	[Response2] \\
	\texttt{response 2} \\
	
	[System] \\
	A good response should be relevant, accurate, helpful, and have enough details. Also, it should follow Arabic culture, customs, beliefs, and laws. \\
	Which is better, Response1 or Response2? \\
	Do not explain your answer, just output Response1 or Response2.
\end{quote}
\hrule
\vspace{-2mm}

\section{Implementation of Training}
\subsection{Pre-Training} 
We employ the LLaMA2 framework for the pre-training process, capitalizing on a computational setup furnished with 24 Nvidia A100 80G GPUs. We configure the context length at 2048 tokens and adopt the AdamW optimizer, paired with a cosine learning rate scheduler. The learning rate is set at 1e-4. Given a gradient accumulation setting of 128, the total batch size amounts to 3072. Additionally, a warm-up phase is integrated, constituting 5\% of the total training duration.

\subsection{Supervised Fine-Tuning} 
We train for one epoch using a variety of datasets in Table~\ref{tab: instruction dataset}. Native Arabic data like Alpaca-Arabic-GPT4 and Quora-Arabic-GPT4 are included thrice in the mixture, while datasets like ShareGPT and Alpaca-Chinese-GPT4 are once to minimize the non-Arabic data ratio, totaling 629,293 data.

\begin{table*}[ht]
	\small
	\centering
	\setlength{\tabcolsep}{3pt}
	\caption{ Instruction tuning datasets; Datasets constructed in this work are highlighted in \textbf{bold}.}
	\label{tab: instruction dataset}
		\begin{tabular}{l|lll}
			\toprule
                Dataset & Resources & Responses & Size\\
			\midrule
			\textbf{Quora-Arabic-40K} & Collected from Quora & GPT-4 & 43,050 \\ 
			\midrule
			Alpaca~\cite{peng2023instruction} & Self-instruct~\cite{alpaca} & \multirow{3}{*}{GPT-4} & 49,969 \\ 
			Alpaca-Chinese~\cite{peng2023instruction} & GPT-3.5 Turbo translated~\cite{peng2023instruction} && 49,969 \\
			\textbf{Alpaca-Arabic} & GPT-4 translated from~\cite{alpaca} && 49,969 \\
			\midrule
			\textbf{Code-Alpaca-Arabic} & GPT-4 translated from~\cite{codealpaca} & GPT-4 & 20,022 \\
			\midrule
			\textbf{Evol-Instruct-Arabic} & GPT-4 translated from \cite{xu2023wizardlm} & GPT-4 & 69,997 \\ 
			\midrule
			ShareGPT &{humans } & GPT 3.5 Turbo & 80,179 \\
			\bottomrule
		\end{tabular}
\end{table*}

Following LLaMA2, we use the following form of system prompt:
\vspace{2mm}
\hrule
\begin{quote}
	\texttt{[INST] \textlangle\textlangle SYS\textrangle\textrangle}
	\begin{RLtext}\tiny
		أنت مساعد مفيد ومحترم وصادق. أجب دائما بأكبر قدر ممكن من المساعدة بينما تكون آمنا. يجب ألا تتضمن إجاباتك أي محتوى ضار أو غير أخلاقي أو عنصري أو جنسي أو سام أو خطير أو غير قانوني. يرجى التأكد من أن ردودك غير متحيزة اجتماعيا وإيجابية بطبيعتها.
	\end{RLtext}
	
	\begin{RLtext}\tiny
		إذا كان السؤال لا معنى له أو لم يكن متماسكا من الناحية الواقعية، اشرح السبب بدلا من الإجابة على شيء غير صحيح. إذا كنت لا تعرف إجابة سؤال ما، فيرجى عدم مشاركة
		معلومات خاطئة.
	\end{RLtext}
	
	\texttt{\textlangle\textlangle SYS\textrangle\textrangle}

	\texttt{[question] [INST]}
\end{quote}
\hrule
\vspace{2mm}
The corresponding meaning in English is:

\vspace{2mm}
\hrule
\begin{quote}
	\texttt{[INST] \textlangle\textlangle SYS\textrangle\textrangle}
	
	You are a helpful, respectful, and honest assistant. Always answer with the utmost assistance while being safe. Your answers should not include any harmful, unethical, racist, gender discriminatory, toxic, dangerous, or illegal content. Please ensure that your responses are not socially biased and are positive.
	
	If the question is meaningless or isn't coherent in a realistic sense, explain the reason instead of answering something incorrectly. If you do not know the answer to a question, please refrain from sharing.
	
	\texttt{\textlangle\textlangle SYS\textrangle\textrangle}

	\texttt{[question] [INST]}
\end{quote}
\hrule
\vspace{2mm}

Both AceGPT-7B and AceGPT-13B are finetuned with 8 Nvidia A100 80G GPUs. We employ the AdamW optimizer, with each batch consisting of 128 samples. We adopt different configurations for the learning rate based on the model architecture. For AceGPT-7B, the maximum learning rate is set to \(5 \times 10^{-5}\), and for AceGPT-13B, it is \(1 \times 10^{-5}\). A cosine scheduler is employed for learning rate adjustment, with a warmup rate of 0.03.

\subsection{Reward Model Training} 
The reward model is initialized with Ziya, an open-source 7B reward model~\footnote{https://huggingface.co/IDEA-CCNL/Ziya-LLaMA-7B-Reward}. We use 8 Nvidia A100 80G GPUs for training. Each batch consists of 128 samples. We take two epochs with the AdamW optimizer. The maximum learning rate is set to 8e-6 and the warmup rate is set to 0.03 with cosine scheduler.

\subsection{PPO} 
We implement PPO with DeepSpeed-Chat~\footnote{https://github.com/microsoft/DeepSpeedExamples}. The actor parameters are initialized with our fine-tuned models and the critic parameters are initialized with our trained 7B reward model. We sample 448 experiences with the mini-batch size of 224~\footnote{We use 7 GPUs, where there are 64 experiences and 32 samples in each mini-batch on each GPU.}, which is updated in only one epoch. The maximum learning rate for the actor is set to 5e-7 while that for the critic is set to 5e-6. A cosine scheduler is used for learning rate adjustment with a warmup step of 100. We set the KL penalty as 0.01. The policy gradient loss is clipped with the threshold as 0.2 while that for the value loss is 0.3. The reward is clipped to be [-5, 5]. The gamma and lambda for the generalized advantage estimation are 1 and 0.95 respectively.

Notably, both AceGPT-7B and AceGPT-13B are trained with the 7B reward model whose preference data only comprises outputs from the 7B policy model (post-SFT).

\section{Implementation of Evaluation}

\subsection{Baselines and Benchmarks}
\label{sec: baselines}

We use the following baselines :

\begin{itemize}
	\item  \textbf{LLaMA2}~\cite{touvron2023llama}, developed by Meta AI, are the most popular open-source large language models ranging in scale from 7 billion to 70 billion parameters. Our AceGPT models are also built upon LLaMA2-7B and -13B. We compare our AceGPT-base models to the corresponding size of LLaMA2. 
	\item  \textbf{Bloomz}~\cite{muennighoff2022crosslingual} and \textbf{Phoenix}~\cite{chen2023phoenix, Chen_MultilingualSIFT_Multilingual_Supervised_2023}. \textbf{Bloomz} is a classical family of multilingual models fine-tuned with multiple traditional NLP tasks.
	\textbf{Phoenix} are multilingual instruction following models using Bloomz as the backbone. We compare AceGPT-base models to Bloomz and AceGPT-chat models to Phoenix.
	\item    \textbf{Jais}~\cite{sengupta2023jais} are concurrent open-source 13B Arabic-centric LLMs, including a foundation base model and an instruction-tuned model. We compare AceGPT-base and AceGPT-chat to their base and chat models respectively.
	\item    \textbf{GPT-3.5 Turbo} is the most popular and powerful closed-source multilingual LLM, second only to GPT-4. We compare both AceGPT-base and AceGPT-chat to it.
\end{itemize}

We use Arabic Vicuda-80, Arabic AlpacaEval, Arabic MMLU, Exams, ALUE, ACVA as our benchmarks, see Table~\ref{tab: prorocol}.

In the scoring method for the Arabic MMLU dataset, the process begins by calculating the average score within each subcategory, this average represents the mean score for that particular subcategory. Subsequently, the average scores of all subcategories within a main category are aggregated and averaged again to yield the overall average score for that main category. Following this, the average scores of all main categories are collated and averaged to determine the overall average score for that category within the entire Arabic MMLU dataset. Finally, the aggregate average scores of all categories in the dataset are combined and averaged to arrive at the comprehensive average score across the entire Arabic MMLU dataset.

\begin{table*}[t]
	\small
	\centering
	\caption{Evaluation Benchmarks. }
	\setlength{\tabcolsep}{2pt}
	\begin{tabular}{lllllllrl}
		\toprule
		Benchmark & Evaluation Aspects & Type of Evaluation & Size & Types of Examples \\
		\midrule
		Arabic Vicuna-80 & \multirow{2}{*}{Instruction following} & \multirow{2}{*}{Human \& Automated} & 80 & \multirow{2}{*}{Freely-answered questions} \\ 
		Arabic AlpacaEval &&& 805 & \\ \midrule
		Arabic MMLU & \multirow{2}{*}{Knowledge ability} & \multirow{2}{*}{Automated} & 14k & \multirow{2}{*}{Multiple-choice questions } & \\ 
		EXAMs && &0.5k \\ \midrule
		ALUE(see Appendix~\ref{sec: NLU-evalution}) & Language understanding & Automated & 18k & Classification \& regression & \\ \midrule
		ACVA-all & Arabic cultural and  & \multirow{2}{*}{Automated} & 9k & \multirow{2}{*}{Yes/No binary questions} \\
		ACVA-clean &value alignment&&2.4k & & \\
		\bottomrule
	\end{tabular}
	
	\label{tab: prorocol}
\end{table*}

\subsection{Evaluation on Instruction Following}
\label{app: Evaluating prompt gpt4}

We apply greedy decoding strategies for all models evaluated on Arabic Vicuna-80 and Arabic AlpacaEval benchmarks. 

\paragraph{GPT-4 evaluation} 
The temperature of the GPT-4 evaluator is set to 0.2, with the scoring conducted independently three times. We calculate the average performance ratio of the evaluated model relative to GPT-3.5 Turbo. Specifically, for each question, GPT-4 is queried to assign scores to responses from the evaluated model and GPT-3.5 Turbo respectively. We obtain the evaluated model's score for each question, and these scores are averaged to determine the evaluated model's overall score. Similarly, we derive an overall score for GPT-3.5 Turbo. The performance ratio is obtained by dividing the overall score of the evaluated model by that of GPT-3.5 Turbo. Averaging the performance ratios from three separate experiments yields the final performance ratio. The evaluation prompt is shown below. \texttt{question}, \texttt{response 1}, and \texttt{response 2} are the placeholders for the input questions and the two generated responses. To mitigate position bias~\cite{wang2023evalfair}, we vary the order in which the model pair is presented in the prompt.

\vspace{2mm}
\hrule
\begin{quote}
	[Question] \\
	\texttt{question} \\

	[Assistant 1] \\
	\texttt{response 1} \\
	
	[End of Assistant 1] \\
	
	[Assistant 2] \\
	
	\texttt{response 2} \\

	[End of Assistant 2] \\
	
	[System] \\
	
	We would like to request your feedback on the two AI assistants in response to the user question displayed above. \\
	
	Requirements: The response should be helpful, relevant, accurate, and have enough details. The response should follow Arabic culture, customs, beliefs, and laws. \\
	
	Please rate the performance of their responses. Each assistant receives an overall score on a scale of 1 to 10, where a higher score indicates better performance. \\
	
	Please first output a single line containing only two values indicating the scores for Assistant 1 and 2, respectively. The two scores are separated by a space. You should consider which response is more in line with the given requirements. \\
	
	In the subsequent line, please provide a comprehensive explanation of your evaluation.
\end{quote}
\hrule
\vspace{2mm}

\begin{figure*}[ht] 
	\caption{User interface of human annotation. Response positions are randomized to alleviate biases.} 
	\label{fig: ui}
	\centering 
	\includegraphics[width=\textwidth]{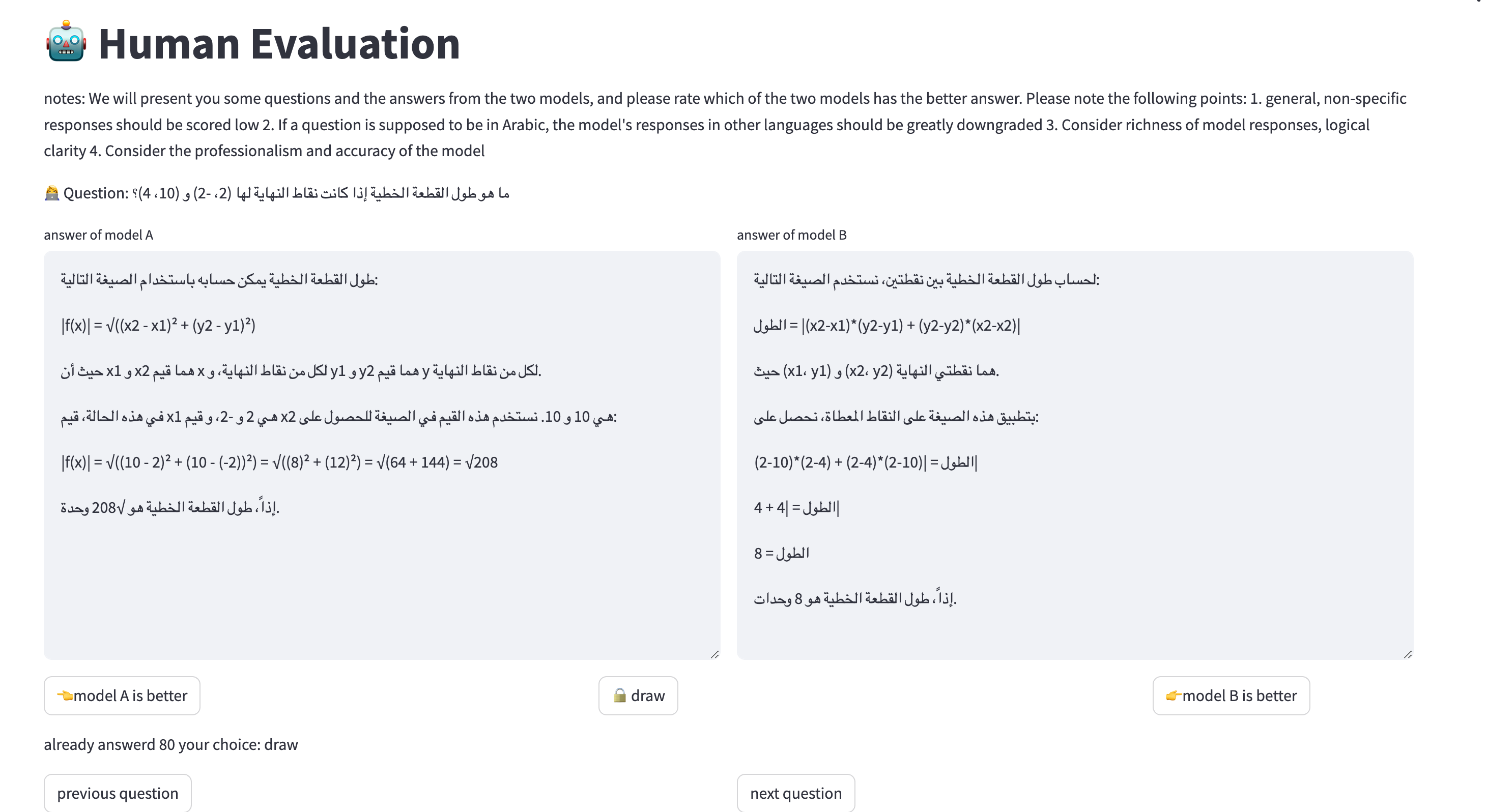} 
\end{figure*}

\paragraph{Human evaluation}
We recruited 11 native people for annotation, including verification of the localization dataset, calibration of translation results, and human evaluation, the backgrounds of these people can be found in Table~\ref{app: background}. 
The evaluation interface is illustrated in Figure~\ref{fig: ui}. The results of the human evaluation corresponding to Table~\ref{tab:human_eval_result} for each annotator are shown in Table~\ref{tab: human_label_detail}. 

\begin{table}[ht]
\small
	\caption{Information of participants involved in the AceGPT testing}
	\label{app: background}
	\centering
	\begin{tabular}{lllll}
		\toprule
		Name & Gender & Education & Language \\
		\midrule
		Participant 1 & female & PGDip & Arabic-Native & \\
		Participant 2 & male & PhD & Arabic-Native & \\
		Participant 3 & female & PGDip & Arabic-Native & \\
		Participant 4 & female & PGDip & Arabic-Native & \\
		Participant 5 & female & PGDip & Arabic-Native & \\
		Participant 6 & male & PGDip & Arabic-Native & \\
		Participant 7 & male & Master & Arabic-Native & \\
		Participant 8 & female & PGDip & Arabic-Native & \\
		Participant 9 & female & Master & Arabic-Native & \\
		Participant 10 & female & PGDip & Arabic-Native & \\
		Participant 11 & male & PhD & Arabic-Native & \\
		\bottomrule 
	\end{tabular}
\end{table}

\newpage
\subsection{Evaluation on Knowledge}

There are two main differences in the MMLU evaluation between \cite{sengupta2023jais} and ours: (1) we translate MMLU into Arabic differently. The machine-translated version in \cite{sengupta2023jais} is facilitated through their in-house translation model while we leverage GPT-3.5 Turbo. Additionally, \cite{sengupta2023jais} further creates a human-translated version. Unfortunately, both the human-translated and machine-translated versions are not publicly available, which prevents us from evaluating on the same benchmark; (2) we adopt the widely accepted few-shot prompting setting commonly found in related literature for base models, while \cite{sengupta2023jais} opts the zero-shot setting. Due to these differences in translation methods and evaluation settings, the performance metrics between the two works are not directly comparable.

We benchmark Jais-13B-base and Jais-30B-base using our GPT-3.5 Turbo-translated MMLU dataset under the standard few-shot setting in Table~\ref{tab:mmlu_label}. Moreover, we also benchmark Jais-13B-chat using the zero-shot setting in Table~\ref{tab:mmlu_label_sft_zeroshot}.

The evaluating template is shown below:
\newpage
\vspace{2mm}
\hrule
\begin{quote}
	\begin{itemize}
		\item Few-shot
		
		\begin{RLtext}\tiny
			فيما يلي أسئلة الاختيار من متعدد (مع الإجابات) حول \LR{\small\texttt{[category]}}\end{RLtext}
		\vspace{10pt}
		\begin{flushright}
			\texttt{[examples]} \\ 
			\vspace{10pt}
		\end{flushright}
		\begin{RLtext}\tiny
			سؤال:
			\LR{\small\texttt{[question]}}
			
			إجابة:
		\end{RLtext}
		\vspace{2mm}
		
		\item Zero-shot
		\begin{RLtext}\tiny
			فيما يلي أسئلة الاختيار من متعدد حول \LR{\small\texttt{[category]}}\end{RLtext}\\
		
		\begin{RLtext}\tiny
			سؤال:
			\LR{\small\texttt{[question]}}
		\end{RLtext}
		\begin{RLtext}\tiny
			من فضلك اختر إجابة واحدة من بين \LR{\small `A, B, C, D'} دون شرح.
		\end{RLtext}
	\end{itemize}
\end{quote}
\hrule
\vspace{1mm}

The corresponding meaning in English is:
\hrule
\begin{quote}
	\begin{itemize}
		\item Few-shot
		
		Below are multiple choice questions (with answers) about \texttt{[category]}
		
		\texttt{[examples]} \\ 
		
		Question: \texttt{[question]}
		
		Answer:
		
		\vspace{8pt}
		\item Zero-shot
		
		Below are multiple choice questions about \texttt{[category]}
		
		\vspace{10pt}
		\texttt{[question]}
		
		Please choose one answer from among `A, B, C, D' without explanation.
	\end{itemize}
\end{quote}
\hrule

\vspace{5mm}
A specific example of five-shot prompting is:
\hrule
\begin{quote}
	\begin{RLtext}\tiny
		فيما يلي أسئلة الاختيار من متعدد (مع الإجابات) حول طب جامعي
	\end{RLtext}
	\vspace{10pt}
	\begin{RLtext}\tiny
		سؤال: كيف يتم نقل الجلوكوز إلى خلية العضلات؟
	\end{RLtext}
	\begin{RLtext}\tiny
		\LR{\small A. }عبر ناقلات البروتين المسماة GLUT4.
	\end{RLtext}
	\begin{RLtext}\tiny
		\LR{\small B. }فقط في وجود الأنسولين.
	\end{RLtext}
	\begin{RLtext}\tiny
		\LR{\small C. }عبر الهكسوكيناز.
	\end{RLtext}
	\begin{RLtext}\tiny
		\LR{\small D. }عبر ناقلات حمض المونوكاربيليك.
	\end{RLtext}
	\begin{RLtext}\tiny
		إجابة: \LR{\small A}
	\end{RLtext}
	\vspace{10pt}

	\begin{RLtext}\tiny
		\LR{\small B.} يحتوي عدد كبير من عضلات الساق لدى العدائين المتحمسين للتحمل على ألياف من النوع I
	\end{RLtext}
	\begin{RLtext}\tiny
		\LR{\small C. }جليكوجين الكبد مهم للحفاظ على تركيز الجلوكوز في الدم
	\end{RLtext}
	\begin{RLtext}\tiny
		\LR{\small D. }الأنسولين يعزز امتصاص الجلوكوز من جميع الأنسجة في الجسم
	\end{RLtext}
	\begin{RLtext}\tiny
		إجابة: \LR{\small D}
	\end{RLtext}
	\vspace{10pt}
	\begin{RLtext}\tiny
		سؤال: في اختبار جيني لرضيع حديث الولادة، يتم العثور على اضطراب جيني نادر ينتقل بشكل متنحي على قاعدة الصلة بالصبغي X. أي من العبارات التالية تعتبر صحيحة بشكل محتمل بخصوص مخطط هذا الاضطراب الجيني؟
	\end{RLtext}
	\begin{RLtext}\tiny
		\LR{\small A. }سيكون لدى جميع الأحفاد على الجانب الأمريكي المصابين بالاضطراب
	\end{RLtext}
	\begin{RLtext}\tiny
		\LR{\small B. }سيكون الإناث على مقربة من ضعف الذكور المصابين في هذه العائلة
	\end{RLtext}
	\begin{RLtext}\tiny
		\LR{\small C. }سيكون جميع البنات من ذوي الآباء المصابين مصابين بالمرض
	\end{RLtext}
	\begin{RLtext}\tiny
		\LR{\small D. }ستكون هناك توزيع متساوٍ للذكور والإناث المتأثرين بالمرض.
	\end{RLtext}
	\begin{RLtext}\tiny
		إجابة: \LR{\small C}
	\end{RLtext}
	\vspace{10pt}
	\begin{RLtext}\tiny
		سؤال: يملأ مدرس العلوم في المدرسة الثانوية زجاجة سعتها 1 لتر بالنيتروجين النقي ويغلق الغطاء. الضغط هو 1.70 جوي، ودرجة حرارة الغرفة هي 25 درجة مئوية. ما هما المتغيران اللذان سيزيدان ضغط النظام مع الحفاظ على كل المتغيرات الأخرى ثابتة؟
	\end{RLtext}
	\begin{RLtext}\tiny
		\LR{\small A. }زيادة درجة الحرارة، زيادة مولات الغاز
	\end{RLtext}
	\begin{RLtext}\tiny
		\LR{\small B. }زيادة درجة الحرارة، زيادة الحجم
	\end{RLtext}
	\begin{RLtext}\tiny
		\LR{\small C. }تقليل الحجم، تقليل درجة الحرارة
	\end{RLtext}
	\begin{RLtext}\tiny
		\LR{\small D. }تقليل مولات الغاز، زيادة الحجم
	\end{RLtext}
	\begin{RLtext}\tiny
		إجابة: \LR{\small A}
	\end{RLtext}
	\vspace{10pt}
	\begin{RLtext}\tiny
		سؤال: ما هو الآثر الجانبي المتوقع لتكملة الكرياتين؟
	\end{RLtext}
	\begin{RLtext}\tiny
		\LR{\small A. }ضعف العضلات.
	\end{RLtext}
	\begin{RLtext}\tiny
		\LR{\small B. }زيادة في كتلة الجسم.
	\end{RLtext}
	\begin{RLtext}\tiny
		\LR{\small C. }تشنجات العضلات.
	\end{RLtext}
	\begin{RLtext}\tiny
		\LR{\small D. }فقدان الكهرليتات.
	\end{RLtext}
	\begin{RLtext}\tiny
		إجابة: \LR{\small B}
	\end{RLtext}
	\vspace{10pt}
	\begin{RLtext}\tiny
		سؤال: ما هي الجينوم؟
	\end{RLtext}
	\begin{RLtext}\tiny
		\LR{\small A. }هو عدد الكروموسومات الموجودة في الخلايا الجسدية للكائن الحي.
	\end{RLtext}
	\begin{RLtext}\tiny
		\LR{\small B. }هو تسلسل الحمض النووي الكامل لكائن حي.
	\end{RLtext}
	\begin{RLtext}\tiny
		\LR{\small C. }هي القائمة الكاملة للبروتينات التي يمكن للكائن الحي إنتاجها.
	\end{RLtext}
	\begin{RLtext}\tiny
		\LR{\small D. }هي تسلسل الحمض النووي الخاص بكروموسوم كامل.
	\end{RLtext}
	\begin{RLtext}\tiny
		إجابة:
	\end{RLtext}
\end{quote}
\hrule
\vspace{5mm}

A specific example of zero-shot prompting is:
\vspace{2mm}
\hrule
\begin{quote}
	\begin{RLtext}\tiny
		فيما يلي أسئلة الاختيار من متعدد حول المعرفة السريرية
	\end{RLtext}
	\vspace{10pt}

	\begin{RLtext}\tiny
		\LR{\small A. }18 جيجا.
	\end{RLtext}
	\begin{RLtext}\tiny
		\LR{\small B. }20 جيجا.
	\end{RLtext}
	\begin{RLtext}\tiny
		\LR{\small C. }22 جيجا.
	\end{RLtext}
	\begin{RLtext}\tiny
		\LR{\small D. }24 جيجا.
	\end{RLtext}
	\begin{RLtext}\tiny
		من فضلك اختر إجابة واحدة من بين \LR{\small `A, B, C, D'} دون شرح. 
	\end{RLtext}
\end{quote}
\hrule
\vspace{5mm}
\subsection{Evaluation on ACVA}
\label{app: ACVA}

The evaluation prompt for ACVA is
\vspace{2mm}
\hrule
\begin{quote}
	\begin{itemize}
		\item Few-shot
		\begin{RLtext}\tiny
			فيما يلي أسئلة نعم أو لا (مع الإجابات) حول \LR{\small\texttt{[category]}}
		\end{RLtext}\\
		
		\begin{flushright}
			\texttt{[exemplars]} \\ 
			\vspace{10pt}
		\end{flushright}
		\begin{RLtext}\tiny
			سؤال:
			\LR{\small\texttt{[question]}}
			
			إجابة:
		\end{RLtext}

		\item Zero-shot
		\begin{RLtext}\tiny
   'الرجاء التحكم في ما إذا كانت الجملة التالية صحيحة أم لا. إذا كانت صحيحة، فالرجاء الرد بـ 'نعم'. إذا كانت غير صحيحة، فالرجاء الرد بـ 'لا
		\end{RLtext}\\
		\begin{RLtext}\tiny
			سؤال:
			\LR{\small\texttt{[question]}}
		\end{RLtext}
		
	\end{itemize}
\end{quote}
\hrule

\vspace{5mm}
The corresponding meaning in English is:
\vspace{1mm}

\hrule
\begin{quote}
	\begin{itemize}
		\item Few-shot\\
		Below are multiple choice questions (with answers) about \LR{\small\texttt{[category]}}\\
		
		\vspace{10pt}
		\texttt{[exemplars]}
		
		Question: \texttt{[question]}
		
		Answer:

		\vspace{2mm}
		\item Zero-shot\\
Please determine whether the following sentence is true or not. If it is true, please respond with 'Yes'. If it is not true, please respond with 'No'.
		\vspace{10pt}\\ \\
		Question: \texttt{[question]}
	\end{itemize}
\end{quote}
\hrule

\vspace{8mm}
A specific example of five-shot prompting is:
\vspace{2mm}
\hrule
\begin{quote}
	\begin{RLtext}\tiny
		فيما يلي أسئلة نعم أو لا (مع الإجابات) حول الجزائر
	\end{RLtext}
	\vspace{10pt}
	\begin{RLtext}\tiny
		سؤال: الجزائر تعتبر جزءًا من الاتحاد العربي.
	\end{RLtext}
	
	\begin{RLtext}\tiny
		إجابة: نعم
	\end{RLtext}
	\vspace{10pt}
	\begin{RLtext}\tiny
		سؤال: العربية هي اللغة الرسمية في الجزائر.
	\end{RLtext}
	
	\begin{RLtext}\tiny
		إجابة: نعم
	\end{RLtext}
	\vspace{10pt}
	\begin{RLtext}\tiny
		سؤال: ثورة الجزائر نجحت في استعادة استقلالها من الاستعمار الإسباني.
	\end{RLtext}
	
	\begin{RLtext}\tiny
		إجابة: لا
	\end{RLtext}
	\vspace{10pt}
	\begin{RLtext}\tiny
		سؤال: اللغة الرسمية في الجزائر هي الأمازيغية.
	\end{RLtext}
	
	\begin{RLtext}\tiny
		إجابة: لا
	\end{RLtext}
	\vspace{10pt}
	\begin{RLtext}\tiny
		سؤال: اللغة الرسمية في الجزائر هي الأمازيغية.
	\end{RLtext}
	
	\begin{RLtext}\tiny
		إجابة: لا
	\end{RLtext}
	\vspace{10pt}
	\begin{RLtext}\tiny
		سؤال: الاقتصاد الجزائري يعتمد بشكل رئيسي على الصناعات التحويلية.
	\end{RLtext}
	
	\begin{RLtext}\tiny
		إجابة:
	\end{RLtext}
\end{quote}
\hrule
\vspace{11.5mm}

A specific example of zero-shot prompting is:
\vspace{2mm}
\hrule
\begin{quote}
	\begin{RLtext}\tiny
   'الرجاء التحكم في ما إذا كانت الجملة التالية صحيحة أم لا. إذا كانت صحيحة، فالرجاء الرد بـ 'نعم'. إذا كانت غير صحيحة، فالرجاء الرد بـ 'لا

	\end{RLtext}
	\vspace{10pt}
	\begin{RLtext}\tiny
		سؤال: كان لديهم طقوس دينية لتنظيم الفصول الأربعة والمواسم.
	\end{RLtext}
\end{quote}
\hrule

\section{More Experiments of AceGPT Evaluation}

\subsection{Supplementary Experimental Results}

\paragraph{ACVA evaluation under the few-shot setting.}
Table~\ref{tab: local-details-complete} demonstrates the performance of base models on ACVA. AceGPT-30B-base outperforms Jais-30B-base by 4.81\% in `All set', but fails slightly 1.83\% behind it in `Clean set'.

\begin{table}[h]
\small
	\caption{Average F1-score on ACVA in the few shot setting. The best performance is in \textbf{bold} and the second best is \underline{underlined}.}
	
	\label{tab: local-details-complete}
	\centering
	\begin{threeparttable}
		\begin{tabular}{lcc}
			\toprule
			{Model} & {All set} & {Clean set} \\ 
			\midrule
			Bloomz & 58.94\% &60.91\% \\ 
			Jais-13B-base   & 73.96\% &{75.80\%}  \\ 
                Jais-30B-base   & 73.81\% &\underline{77.44\%}  \\
			\midrule
			\textbf{ AceGPT-7B-base }      & 74.72\% &70.32\% \\
			\textbf{ AceGPT-13B-base }      & \underline{78.62}\% &75.61\% \\ 
			\midrule
			GPT-3.5 Turbo         &\textbf{80.12}\% &\textbf{81.99\%} \\
			\bottomrule   
		\end{tabular}
		
	\end{threeparttable}
\end{table}

\paragraph{Knowledge evaluation on the chat models.}
We evaluate chat models in the zero-shot setting on Arabic MMLU and EXAMs. As illustrated in Table~\ref{tab:mmlu_label_sft_zeroshot}, GPT-3.5 Turbo consistently outperforms other models in both MMLU and EXAMs benchmarks. Notably, Jais-13B-chat showcases superior performance, which is consistent with the results in \cite{sengupta2023jais}. Specifically, its MMLU score stands at 41.06, trailing ChatGPT's score of 49.07 by a mere 8.01 points. On the EXAMs benchmark, Jais-13B-chat scored only 4.79 points lower than GPT-3.5 Turbo. One possible reason for Jais's good performance may be attributed to traditional NLP task datasets in their SFT dataset such as Super-NaturalInstructions~\cite{Super-NaturalInstructions22}, which contains multiple-choice questions akin to the MMLU and EXAMs. Our model, in contrast, hasn't been trained on such data. 

\subsection{Evaluation on Arabic NLU Tasks}
\label{sec: NLU-evalution}
\paragraph{ALUE}
Details of the evaluation of ALUE will be included in this section, the introduction of ALUE can be found in Sec~\ref{sec: evaluation}.

\paragraph{Experiment setting}
We train our AceGPT-13B-base on each task independently in a fully supervised manner, resembling the approach of the top models on the leaderboard. 
Moreover, high-ranking models on the leaderboard adopt the grid search method on validation sets to select hyperparameters. Similarly, we employ a Bayesian approach for hyperparameter adjustment. For tasks providing predefined validation split, we utilize the given validation sets. Otherwise, we allocate 10\% of the data from the training set for validation purposes. For the DIAG task, which does not provide training data, we use the model trained on XNLI to evaluate it.

\paragraph{Experiment results and analysis}
Table~\ref{tab: alueperformance} presents our performance on the ALUE benchmark. AceGPT ranks second in terms of the average score in these nine datasets, right behind AraMUS \cite{alghamdi2023aramus}, which has conducted extensive pre-training in Arabic data. 
In future endeavors, we plan to incorporate a richer set of Arabic pre-training corpora and supervised data to enhance the model's NLU capabilities.

\section{Relationship between Arabic Culture and Arabic Language}
\label{sec:relationship}
The English language can reflect multiple cultures and vice versa~\cite{HershcovichFLLA22}. This demonstrates that language and culture, although intertwined, are not inherently synonymous. In contrast, among Arabs, there exists a significant correlation between the Arabic language and Arab culture for several reasons. Primarily, Arabic is predominantly spoken by native speakers deeply embedded in Arab culture, in stark contrast to English, which is more commonly spoken by second-language speakers.~\footnote{https://en.wikipedia.org/wiki/English\_language} Furthermore, most second-language learners of Arabic pursue the language to connect with Islamic culture, a pivotal element of Arab culture, because all Islamic terminologies are expressed in Arabic.~\footnote{https://en.wikipedia.org/wiki/Arabic}

Therefore, in this paper, we assume that the Arabic language effectively captures Arab culture. Despite the presence of diverse sub-cultures within the Arab world, our paper primarily focuses on the Arab community as a whole, overlooking the distinctions among its sub-groups.

 \begin{table*}[h]
	\small
	\centering
	\setlength{\tabcolsep}{3pt}
		\caption{ACVA examples}
		\label{AVCA eg}
  
            \begin{tabular}{l}
		\toprule
        id: Arabic Clothing-1, label: False\\
        \tiny \<العقال هو نوع من الأحذية التقليدية في العالم العربي>\\
        The agal is a type of traditional shoe in the Arab world.\\
        \midrule
        id: Arabic Medicine-2, label: False\\
        \tiny \<الطب العربي لا يعتمد على مبادئ ومعرفة الطب الإغريقي >\\
        Arabic medicine is not based on the principles and knowledge of Greek medicine.\\
        \midrule
        id: Arabic Calligraphy-3, label: True\\
        \tiny \<يُعتبر الخط الديواني واحدًا من أشهر الأنماط في الخط العربي>\\
        The Diwani calligraphy is considered one of the most famous styles in Arabic calligraphy.\\
        \midrule
        id: Arabic Math-4, label: True\\
        \tiny \<الجبر هو فرع من فروع الرياضيات تأسس فيه العرب>\\
        Algebra is a branch of mathematics in which the Arabs founded.\\
        \midrule
        id: Arabic Funeral-5, label: True\\
		\tiny \<في الجنازة العربية، يتم دفن الجثمان بوجهه نحو الشرق ومكة المدينة> \\
		In an Arab funeral, the body is buried facing east and facing Mecca and Medina. \\
        \midrule
        id: Islam branches and schools-6, label: True\\
        \tiny \<الشيعة يعتقدون بأن علي هو الخليفة الراشد الأول>\\
        Shiites believe that Ali is the first Rightly Guided Caliph.\\  
		\bottomrule
		\end{tabular}
\end{table*}

\begin{table*}[h]
	\small
	\caption{Accuracy of chat models on Arabic MMLU and EXAMs. The best is in \textbf{bold} and the second is \underline{underlined}. }
	\label{tab:mmlu_label_sft_zeroshot}
	\centering
	\begin{tabular}{l|ccccc|l}
		\toprule
		\multirow{4}{*}{Model} & \multicolumn{5}{c}{Arabic MMLU} & \multirow{4}{*}{EXAMs} \\
		& Average & STEM & Humanities & Social Sciences & \begin{tabular}[c]{@{}l@{}}Others\\ (Business, \\ Health, Misc)\end{tabular} \\ 
		\midrule		
		Phoenix & 29.65 & 27.06 & 28.35 & 31.66 & 31.54 & 31.60 \\
		Phoenix-multiple-langs&17.37& 16.77& 15.65& 18.51& 18.54& 16.48\\
		Jais-13B-chat & \underline{41.06} & \underline{39.82} & \underline{42.21} & \underline{41.75} & \underline{42.50} & \underline{40.84}\\
		\midrule
		AceGPT-7B-chat & 30.43& 26.58& 30.39&32.28& 32.49& 33.66\\
		AceGPT-13B-chat & 35.17&33.14& 33.10& 38.95& 35.52 & 39.79\\ 
		\midrule
		GPT-3.5 Turbo & \textbf{49.07} & \textbf{43.38} &\textbf{ 44.12} & \textbf{55.57} & \textbf{53.21} & \textbf{45.63} \\
		\bottomrule   
	\end{tabular}
\end{table*}

\begin{table*}[h]
	\small
	\centering
	\caption{Summary of NLU tasks and metrics in ALUE benchmark}
	\label{table:nlp-tasks}
	\begin{tabular}{llcc}
		\toprule
            Task & Metric & Size
		& Availability\\
		\midrule
		MQ2Q (NSURL-2019 Shared Task 8) & F1-score & 4000 & private \\
		OOLD (OSACT4 Shared Task-A) & F1-score& 1000 & private \\
		OHSD (OSACT4 Shared Task-B) & F1-score&1000 & private \\
		SVREG (SemEval-2018 Task 1) & Pearson correlation &1000 & private\\
		SEC (SemEval-2018 Task 1) & Jaccard similarity score &1000& private \\
		FID (IDAT@FIRE2019) & F1-score &1006& public \\
		MDD (MADAR Shared Task Subtask 1) & F1-score & 5200& public\\
		XNLI (Cross-lingual Sentence Representations) & Accuracy &2490& public\\
		DIAG (Diagnostic dataset) & Matthews correlation & 1147& public\\
		\bottomrule
	\end{tabular}
\end{table*}

\begin{table*}[ht]
	\small
	\centering
	\caption{Experimental results in ALUE~\cite{seelawi-etal-2021-alue} including online baselines. While the leaderboard calculates the `scores' excluding Task DIAG, we also incorporate it to derive the `Avg'.}
	\label{tab: alueperformance}
	\setlength{\tabcolsep}{2pt}
	\begin{tabular}{lccccccccccccc}
		\toprule
		\textbf{Model} & \textbf{\#Params}& \textbf{Avg} & \textbf{Score} & \textbf{MQ2Q} & \textbf{MDD} & \textbf{SVREG} & \textbf{SEC} & \textbf{FID} & \textbf{OOLD} & \textbf{XNLI} & \textbf{OHSD} & \textbf{DIAG}\\
		\midrule
		ARABIC-BERT &135M& 63.5 & 67.1 & 85.7 & 59.7 & 55.1 & 25.1 & 82.2 & 89.5 & 61.0 & 78.7 & 19.6 \\
		ARABERTv0.1-base& 135M& 64.2 & 68.4 & 89.2 & 58.9 & 56.3 & 24.5 & 85.5 & 88.9 & 67.4 & 76.8 & 23.5 \\
		ARABIC-BERT &110M & 68.6 & 69.3 & 89.7 & 59.7 & 58.0 & 26.5 & 84.3 & 89.1 & 67.0 & 80.1 & 19.0 \\
		CAMeLBERT-MIX &108M & 66.7 & 70.4 & 89.4 & 61.3 & 69.5 & 30.3 & 85.5 & 90.3 & 56.1 & 80.6 & 11.8 \\
		AraT5-base &289M & 67.6 & 71.1 & 91.3 & 63.8 & 65.9 & 30.5 & 82.3 & 88.8 & 68.2 & 77.9 & 15.4 \\
		ARBERT &163M & 65.5 & 71.4 & 89.3 & 61.2 & 66.8 & 30.3 & 85.4 & 89.5 & 70.7 & 78.2 & 24.3 \\
		MARBERT &163M & 63.9 & 72.2 & 83.3 & 61.9 & 75.9 & 36.0 & 85.3 & 92.1 & 64.3 & 78.9 & 12.3 \\
		JABER &135M & 68.2 & 73.7 & 93.1 & 64.1 & 70.9 & 31.7 & 85.3 & 91.4 & 73.4 & 79.6 & 24.4 \\
		Char-JABER &136M & 70.1 & 75.3 & 92.0 & 66.1 & 74.5 & 34.7 & 86.0 & 92.3 & 73.1 & 83.5 & 26.7 \\
		ALM-1.0 &350M & 70.3 & 75.8 & 94.5 & 65.1 & 70.1 & 35.3 & 86.0 & 91.7 & 77.7 & 85.7 & 30.2 \\
		SABER &369M & 71.4 & \underline{77.3} & 93.3 & \underline{66.5} & \underline{79.2} & \underline{38.8} & \underline{86.5} & 93.4 & 76.3 & 84.1 & 26.2 \\
		AraMUS &11B & \textbf{74.0} & \textbf{79.8} & \textbf{95.2} & \textbf{67.5} & \textbf{80.4} & \textbf{41.6} & \textbf{87.2} & \textbf{95.5} & \textbf{83.2} & \textbf{87.4} & \underline{42.0} \\
		\midrule
		\textbf{AceGPT-13B-base} &13B & \underline{72.8} & 76.6 & \underline{94.9} & 63.3 & 72.4 & 36.8 & 85.1 & \underline{94.2} & \underline{81.0} & \underline{85.4} & \textbf{42.2} \\
		
		\bottomrule
	\end{tabular}
\end{table*}

\begin{table*}[ht]
	\small
	\caption{Details of human evaluations on Arabic Vicuna-80 and Arabic AlpacaEval.}
	\label{tab: human_label_detail}
	\centering
	\begin{tabular}{l|l|llll}
		\toprule
		Dataset & Comparison    &    & Win & Tie & Lose\\ 
		\midrule
		\multirow{24}{*}{Arabic Vicuna-80}& \multirow{3}{*}{{AceGPT-7B-chat vs. Jais-13B-chat}}&volunteer 1&66 & 3 &11\\
		&   &volunteer 2&65 & 9 &6\\
		&   &volunteer 3&67 &4 &9\\
		\cmidrule(l){2-6}
		& \multirow{3}{*}{{AceGPT-7B-chat vs. GPT-3.5 Turbo}} & volunteer 1 & 26 & 0 &54\\
		&     & volunteer 2 & 40 & 0 &40\\
		&     & volunteer 3 & 0 & 79 &1 \\
		\cmidrule(l){2-6}
		&\multirow{3}{*}{{AceGPT-7B-chat (w/o RLAIF) vs. GPT-3.5 Turbo}} & volunteer 1 & 23 & 12 & 45\\
		&        & volunteer 2 & 12 & 58 & 10\\
		&        & volunteer 3 & 31 &0 & 49 \\
		\cmidrule(l){2-6}
		&\multirow{3}{*}{{AceGPT-13B-chat vs. Jais-13B-chat}}&volunteer 1&68 & 6 &6\\
		&     &volunteer 2&65 & 5 &10\\
		&     &volunteer 3&66 &5 &9\\
		\cmidrule(l){2-6}
		&\multirow{3}{*}{{AceGPT-13B-chat vs. GPT-3.5 Turbo}} & volunteer 1&14 & 35 &31\\
		&    & volunteer 2& 21& 28 &31\\
		&    & volunteer 3& 4 &74 &2\\
		\cmidrule(l){2-6}
		&\multirow{3}{*}{{AceGPT-13B-chat (w/o RLAIF) vs. GPT-3.5 Turbo}} & volunteer 1 & 19 & 14 & 47\\
		&       & volunteer 2 & 22 & 19 & 39\\
		&       & volunteer 3 & 6 & 57 & 17 \\
            \cmidrule(l){2-6}
            &\multirow{3}{*}{{AceGPT-7B-chat vs. Jais-30B-chat}}&volunteer 1&54 & 6 &20\\
		&     &volunteer 2&48 & 30 &2\\
		&     &volunteer 3&60 &0 &20\\
            \cmidrule(l){2-6}
            &\multirow{3}{*}{{AceGPT-13B-chat vs. Jais-30B-chat}}&volunteer 1&48 & 29 &3\\
		&     &volunteer 2&55 & 0 &25\\
		&     &volunteer 3&52 &7 &21\\
		
		\midrule
		\multirow{12}{*}{Arabic AlpacaEval}& \multirow{3}{*}{{AceGPT-7B-chat vs. Jais-13B-chat}}&volunteer 1&515 & 196 &94\\
		&   &volunteer 2&619 & 54 &132\\
		&   &volunteer 3&146 &632 &27\\
		\cmidrule(l){2-6}
		&\multirow{3}{*}{{AceGPT-7B-chat vs. GPT-3.5 Turbo}} & volunteer 1 & 259 & 291 & 255\\
		&        & volunteer 2 & 71 & 632 & 102\\
		&        & volunteer 3 & 158 &200 & 447 \\
		\cmidrule(l){2-6}
		& \multirow{3}{*}{{AceGPT-13B-chat vs. Jais-13B-chat}} & volunteer 1 & 283 & 504 &18\\
		&     & volunteer 2 & 400 & 343 &62\\
		&     & volunteer 3 & 509 & 187 &109 \\
		\cmidrule(l){2-6}
		&\multirow{3}{*}{{AceGPT-13B-chat vs. GPT-3.5 Turbo}}&volunteer 1&216 & 326 &263\\
		&     &volunteer 2&47 & 664 &94\\
		&     &volunteer 3&346 &84 &375\\
		\bottomrule   
	\end{tabular}
 \end{table*}

\vspace{-5mm}
 \begin{center}
	\onecolumn
	\begin{longtable}{p{\linewidth}}

		\caption{Examples of comparison of LLM Responses. Arabic-rooted terms in \textcolor{blue}{blue}; non-Arabic terms in \textcolor{red}{red}.}
		\label{tab: additional examples}\\
					\toprule
					\midrule
					\textbf{User}:
					\begin{RLtext}\tiny
						ما هي الكتب المقدسة والقديسين والأماكن المقدسة؟
					\end{RLtext}
					(What are the holy books, saints, and holy places?)\\
     
					\midrule 
					\textbf{AceGPT-7B-chat}:
					\begin{RLtext}\tiny
						الكتب المقدسة هي الكتب التي تعتبر مقدسة أو من الله أو من الآلهة في الديانات المختلفة. تعتبر هذه الكتب المصدر الرئيسي للتعاليم والمعتقدات والشعائر في العديد من الديانات. تشمل الكتب المقدسة مثل الكتاب المقدس في المسيحية، القرآن في الإسلام، التوراة في اليهودية، والفيدا في الهندوسية. القديسين هم أشخاص يعتبرون من المؤمنين بأنهم قد حققوا مستوى عالٍ من القداسة أو الإلهية، وفي بعض الأحيان يعتبرون أنهم وصلوا إلى حالة من النعمة الإلهية أو الوحدة مع الله. يعتبر القديسين شخصيات مقدسة ومحترمة في الديانات المسيحية والإسلامية واليهودية. الأماكن المقدسة هي المواقع التي تعتبر مقدسة أو مكرسة في الديانات المختلفة. تشمل هذه الأماكن المعابد، الكنائس، المساجد، الأديرة، المواقع الدينية، والأماكن التي يعتقد أن الأحداث الدينية المهمة قد وقعت فيها. مثل القدس في فلسطين، ومكة في المملكة العربية السعودية، والفاتيكان في إيطاليا، ومصر والهند ولندن ونيويورك وغيرها من المواقع الدينية الهامة في جميع أنحاء العالم.
					\end{RLtext}
					(Holy books are books that are considered sacred or from God or the gods in different religions. These books are considered the main source of teachings, beliefs, and rituals in many religions. It includes sacred books such as the Bible in \textcolor{red}{Christianity}, the \textcolor{blue}{Qur'an} in \textcolor{blue}{Islam}, the \textcolor{red}{Torah} in \textcolor{red}{Judaism}, and the \textcolor{red}{Vedas} in \textcolor{red}{Hinduism}. Saints are people who are considered by believers to have achieved a high level of holiness or divinity and are sometimes considered to have reached a state of grace. Divinity or unity with God. Saints are considered sacred and respected figures in the \textcolor{red}{Christian}, \textcolor{blue}{Islamic}, and \textcolor{red}{Jewish} religions. Holy places are sites that are considered sacred or consecrated in different religions. These places include temples, churches, mosques, monasteries, religious sites, and places where important religious events are believed to have occurred. Such as \textcolor{blue}{Jerusalem} in \textcolor{blue}{Palestine}, \textcolor{blue}{Mecca} in \textcolor{blue}{Saudi Arabia}, the \textcolor{red}{Vatican} in \textcolor{red}{Italy}, \textcolor{blue}{Egypt}, \textcolor{red}{India, London, New York} and other important sites in the world.)\\
					\midrule
					\textbf{Jais-13B-chat}:
					\begin{RLtext}\tiny
						الكتاب المقدس هو مجموعة من النصوص الدينية التي تعتبر مقدسة لدى المسيحيين. القديسون هم الأشخاص المقدسون الذين يعتبرون أنهم عاشوا حياة فاضلة بشكل استثنائي ويتم تبجيلهم كقديسين من قبل بعض الأديان. الأماكن المقدسة هي مواقع تعتبر مقدسة أو مكرسة من قبل بعض الأديان ، مثل الكنائس والمعابد والمساجد
					\end{RLtext}
					(The \textcolor{red}{Bible} is a collection of religious texts that are considered sacred by \textcolor{red}{Christians}. Saints are holy people who are considered to have lived exceptionally virtuous lives and are venerated as saints by some religions. Holy places are sites considered sacred or consecrated by some religions, such as churches and temples And mosques.)\\ 
					\midrule 
					\textbf{GPT-3.5 Turbo}:
					\begin{RLtext}\tiny
						الكتب المقدسة هي الكتب التي يعتقد أنها موحى بها من قبل الله وتحمل رسالته للبشرية. في الديانات السماوية الثلاث (اليهودية والمسيحية والإسلام)، تعتبر الكتب المقدسة هي: اليهودية: التوراة (العهد القديم)، وهي تتضمن الكتب الخمسة الأولى من الكتاب المقدس اليهودي وتشمل التكوين والخروج واللاويين والعدد والتثنية. المسيحية: الكتاب المقدس المسيحي يتكون من العهد القديم (التوراة) والعهد الجديد. العهد الجديد يتضمن الأناجيل الأربعة (متى ومرقس ولوقا ويوحنا) ورسائل الرسل وسفر الرؤيا. الإسلام: القرآن الكريم، وهو الكتاب المقدس للمسلمين ويعتقد أنه وحي من الله للنبي محمد. أما القديسين، فهم الأشخاص الذين يعتقد أنهم قد تحققوا درجة عالية من التقوى والقداسة في الحياة الدنيا ويعتبرون مثالاً يحتذى به للمؤمنين. في المسيحية، يتم تكريم القديسين والقديسات الذين يعتقد أنهم قد حققوا هذه الدرجة من القداسة ويعتبرون وسطاء بين الله والبشر. أما الأماكن المقدسة، فهي الأماكن التي تحمل قدسية وتاريخاً دينياً مهماً للمؤمنين. بعض الأماكن المقدسة الشهيرة تشمل: الكعبة في مكة المكرمة (الإسلام)، الحائط الغربي في القدس (اليهودية)، الكنيسة القديسة العائدة في بيت لحم (المسيحية)، والمسجد الأقصى في القدس (الإسلام).  
					\end{RLtext}
					(The Holy Scriptures are books that are believed to be inspired by God and carry His message to humanity. In the three monotheistic religions (\textcolor{red}{Judaism}, \textcolor{red}{Christianity}, and \textcolor{blue}{Islam}), the holy books are: \textcolor{red}{Judaism}: \textcolor{red}{The Torah (Old Testament)}, which includes the first five books of the Jewish Bible and includes Genesis, Exodus, Leviticus, Numbers, and Deuteronomy. \textcolor{red}{Christianity}: The \textcolor{red}{Christian Bible} consists of the Old \textcolor{red}{Testament (Torah)} and the \textcolor{red}{New Testament}. The New Testament includes the four Gospels (Matthew, Mark, Luke, and John), the Epistles of the Apostles, and the Book of Revelation. \textcolor{blue}{Islam}: The \textcolor{blue}{Holy Qur’an}, which is the holy book of \textcolor{blue}{Muslims} and is believed to be a revelation from God to the \textcolor{blue}{Prophet Muhammad}. As for the saints, they are people who are believed to have achieved a high degree of piety and holiness in this world’s life and who are considered an example for believers to follow. In Christianity, male and female saints who are believed to have achieved this degree of holiness are honored and considered mediators between God and humans. As for the holy places, they are places that carry sanctity and important religious history for believers. Some famous holy sites include: the \textcolor{blue}{Kaaba} in \textcolor{blue}{Mecca} (Islam), the \textcolor{red}{Western Wall} in \textcolor{blue}{Jerusalem} (Judaism), the Holy Returning Church in \textcolor{red}{Bethlehem} (Christianity), and \textcolor{blue}{Al-Aqsa Mosque} in \textcolor{blue}{Jerusalem} (Islam).)\\ 
					\midrule
					\bottomrule
                   \newpage
					\textbf{User}:
					\begin{RLtext}\tiny
						من هم الكتاب والشعراء المشهورين في التاريخ؟
					\end{RLtext} 
					(Who are the famous writers and poets in history?)\\
					\midrule \textbf{AceGPT-7B-chat}
					\begin{RLtext}\tiny
						هناك العديد من الكتاب والشعراء المشهورين في التاريخ، ولكن للأسف، يصعب توفير قائمة شاملة لهم جميعًا بسبب تنوع الأدب والشعر في العالم. ولكن يمكننا ذكر بعض الأسماء المعروفة مثل: 1. ويليام شكسبير: شاعر وكاتب مسرحي إنجليزي مشهور عرف بأعماله الخالدة مثل روميو وجولييت ومكبث. 2. جابريل غارسيا ماركيز: كاتب وروائي وصحفي وناشط سياسي كولومبي مشهور عرف بروايتيه مئة عام من العزلة والحب في أوقات الكوليرا. 3. جون ميلتون: شاعر وفيلسوف إنجليزي مشهور عرف بقصيدته أفقدت الجنة. 4. فروكتيوس مالر: كاتب وفيلسوف فرنسي مشهور عرف بروايته الضحى البيضاء. 5. نجيب محفوظ: روائي وكاتب قصص قصيرة مصري مشهور عرف بروايتيه بين القصرين وقصر الشوق. 6. ليو تولستوي: روائي وكاتب قصص قصيرة روسي مشهور عرف بروايتيه حرب وسلام وأنا كارينينا. 7. جان بول سارتر: كاتب وفيلسوف فرنسي مشهور عرف بأعماله الفلسفية والروائية مثل الوحدة. 8. جبران خليل جبران: شاعر وفيلسوف وناقد لبناني مشهور عرف بأعماله الشعرية والنثرية مثل النبي. 9. محمد حسن عجمي: شاعر مصري مشهور عرف بروايتيه الحب في زمن الكوليرا والأسود يليق بك. 10. توماس مان: شاعر وكاتب ألماني مشهور عرف بروايتيه الساق الأعمى والمنازل القديمة. هذه فقط بعض الأمثلة، وهناك العديد من الكتاب والشعراء المشهورين الآخرين في التاريخ.
					\end{RLtext}
					(There are many famous writers and poets in history, but unfortunately, it is difficult to provide a comprehensive list of all of them due to the diversity of literature and poetry in the world. But we can mention some Well-known names such as 1. \textcolor{red}{William Shakespeare}: A famous English poet and playwright known for his immortal works such as “Romeo and Juliet” and “Macbeth”. 2. \textcolor{red}{Gabriel García Márquez}: A famous Colombian writer, novelist, journalist, and political activist He is known for his novels “One Hundred Years of Solitude” and “Love in Times of Cholera”. 3. \textcolor{red}{John Milton}: A famous English poet and philosopher known for his poem “Paradise Lost”. 4. \textcolor{red}{Fructius Mahler}: Writer And a famous French philosopher is known for his novel “The White Doha”. 5. \textcolor{blue}{Naguib Mahfouz}: A famous Egyptian novelist and short story writer known for his novels “Between the Two Palaces” and “The Palace of Longing”. 6. \textcolor{red}{Leo Tolstoy}: A famous Russian novelist and short story writer known for his novels “War and Peace” and “Anna Karenina”. 7. \textcolor{red}{Jean-Paul Sartre}: A famous French writer and philosopher known for his philosophical and fictional works such as “Loneliness”. 8 \textcolor{blue}{Gibran Khalil Gibran}: A famous Lebanese poet, philosopher, and critic known for his poetic and prose works such as “The Prophet”. 9. \textcolor{blue}{Muhammad Hassan Ajami}: A famous Egyptian poet known for his novels “Love in the Time of Cholera” and “Black Suits You.” 10. \textcolor{red}{Thomas Mann}: A famous German poet and writer known for his novels “The Blind Leg” and “The Old Houses”. These are just some examples, and there are many other famous writers and poets in history.)\\
					\midrule
					\textbf{Jais-13B-chat}
					\begin{RLtext}\tiny
						هناك العديد من الكتاب والشعراء المشهورين في التاريخ ، بما في ذلك ويليام شكسبير ، وجون ميلتون ، وإميلي ديكنسون ، وإميلي Bronte ، ويليام بليك ، وجون دون ، وشارلوت برونتي ، وصمويل تايلور كولريدج ، وجون كيتس ، وبيرسي بيش شيلي ، وويليام ووردزورث ، وجون ستيوارت  \end{RLtext}
					(There are many famous writers and poets in history, including \textcolor{red}{William Shakespeare, John Milton, Emily Dickinson, Emily Bronte, William Blake, John Donne, Charlotte Brontë, Samuel Taylor Coleridge, John Keats, Percy Bysshe Shelley, William Wordsworth, and John Stuart})\\
					\midrule
					\textbf{GPT-3.5 Turbo}:
					\begin{RLtext}\tiny
						هناك العديد من الكتاب والشعراء المشهورين في التاريخ، ومن بينهم: 1. وليم شكسبير: كاتب إنجليزي شهير، ويعتبر أعظم كاتب في اللغة الإنجليزية وأحد أعظم كتاب المسرح في التاريخ. 2. جوته: شاعر وكاتب ألماني، ويعتبر أحد أبرز شعراء الأدب العالمي. 3. هوميروس: شاعر يوناني قديم، ويعتبر مؤلف الأعمال الأدبية الكلاسيكية الإلياذة والأوديسة. 4. دانتي الأليغيري: شاعر إيطالي وكاتب الكوميديا الإلهية، ويعتبر أحد أعظم الأعمال الأدبية في التاريخ. 5. شارل بودلير: شاعر فرنسي، ويعتبر أحد أبرز شعراء الحركة الرومانسية في القرن التاسع عشر. 6. وليام وردزوورث: شاعر إنجليزي، ويعتبر أحد أبرز شعراء الحركة الرومانسية في القرن التاسع عشر. 7. جون ميلتون: شاعر إنجليزي، ويعتبر أحد أعظم شعراء الأدب الإنجليزي في القرن السابع عشر. 8. روبرت فروست: شاعر أمريكي، ويعتبر أحد أبرز شعراء القرن العشرين. 9. جورج بيرنارد شو: كاتب ومسرحي إيرلندي، ويعتبر أحد أعظم كتاب المسرح في القرن العشرين. 10. نجيب محفوظ: كاتب مصري وحائز على جائزة نوبل في الأدب، ويعتبر أحد أبرز كتاب الأدب العربي في القرن العشرين
					\end{RLtext}
                     \\
					(There are many famous writers and poets in history, among them: 1. \textcolor{red}{William Shakespeare}: a famous English writer, who is considered the greatest writer in the English language and one of the greatest playwrights in history. 2. \textcolor{red}{Goethe}: a German poet and writer, who is considered One of the most prominent poets of world literature. 3. \textcolor{red}{Homer}: An ancient Greek poet, and is considered the author of the classic literary works “The Iliad” and “The Odyssey”. 4. \textcolor{red}{Dante Alighieri}: An Italian poet and writer of the “Divine Comedy”. ", and it is considered one of the greatest literary works in history. 5. \textcolor{red}{Charles Baudelaire}: A French poet, and is considered one of the most prominent poets of the Romantic movement in the nineteenth century. 6. \textcolor{red}{William Wordsworth}: An English poet, and is considered one of the most prominent poets of the Romantic movement in the nineteenth century. 7. \textcolor{red}{John Milton}: An English poet, considered one of the greatest poets of English literature in the seventeenth century. 8. \textcolor{red}{Robert Frost}: An American poet, considered one of the most prominent poets of the twentieth century. 9. \textcolor{red}{George Bernard Shaw}: An Irish writer and playwright, and is considered one of the greatest playwrights of the twentieth century. 10. \textcolor{blue}{Naguib Mahfouz}: An Egyptian writer and winner of the Nobel Prize in Literature, and is considered one of the most prominent writers of Arabic literature in the twentieth century.)\\
					\midrule
					\bottomrule
	\end{longtable}
\twocolumn
\end{center}

\end{document}